%% file: arxiv.tex
\documentclass[11pt]{article}

\usepackage[preprint]{acl}

\usepackage{times}
\usepackage{latexsym}
\usepackage{amsmath}
\usepackage{amssymb}
\usepackage{mathtools}
\usepackage{enumitem}
\usepackage{amsthm}

\usepackage{amsmath}
\usepackage[T1]{fontenc}
\usepackage{todonotes}

\usepackage{mdframed}

\usepackage[utf8]{inputenc}
\usepackage{tabularray}
\usepackage{microtype}
\usepackage{xcolor}
\usepackage{booktabs}
\usepackage{hyperref}
\usepackage{cleveref}
\usepackage[table]{xcolor}
\newmdenv[
  linecolor=orange!80!black,
  linewidth=3pt,
  topline=false, bottomline=false, rightline=false,
  innerleftmargin=8pt, innerrightmargin=4pt,
  innertopmargin=3pt, innerbottommargin=3pt,
  skipabove=4pt, skipbelow=4pt,
]{todoblockenv}

\definecolor{paperBg}{HTML}{FBFAF7}        
\definecolor{paperBorder}{HTML}{C9C5BB}    
\definecolor{paperRare}{HTML}{E6550D}      
\definecolor{paperRareDark}{HTML}{A6360A}  
\definecolor{paperTARQ}{HTML}{D62728}      
\definecolor{paperGood}{HTML}{2CA02C}      
\definecolor{paperBad}{HTML}{888888}       
\definecolor{paperRef}{HTML}{555555}       

\newmdenv[
  linecolor=paperBorder,
  linewidth=0.5pt,
  backgroundcolor=paperBg,
  roundcorner=3pt,
  innerleftmargin=7pt, innerrightmargin=7pt,
  innertopmargin=5pt, innerbottommargin=5pt,
  skipabove=5pt, skipbelow=5pt,
  nobreak=true,
]{rwexamplebox}
\newcommand{\rwOK}{{\color{paperGood}\cmark}}
\newcommand{\rwNO}{{\color{paperBad}\xmark}}
\newcommand{\rwWord}[1]{{\color{paperRareDark}\textbf{#1}}}
\newcommand{\rwTARQ}{{\color{paperTARQ}\textbf{\TARQ}}}
\newcommand{\rwRef}{{\color{paperRef}\textit{Ref:}}}
\usepackage{inconsolata}

\usepackage{multirow}
\usepackage{pifont}
\usepackage{subcaption}
\newcommand{\cmark}{\ding{51}}
\newcommand{\xmark}{\ding{55}}
\usepackage[ruled,vlined]{algorithm2e}
\crefname{algocf}{Algorithm}{Algorithms}
\Crefname{algocf}{Algorithm}{Algorithms}
\usepackage{graphicx}
\usepackage{placeins}    
\usepackage{float}       
\usepackage{caption}
\makeatletter
\AtBeginDocument{%
  \setlength{\@fptop}{0pt}%
  \setlength{\@fpsep}{8pt plus 0pt}%
  \setlength{\@fpbot}{0pt plus 1fil}%
  \setlength{\@dblfptop}{0pt}%
  \setlength{\@dblfpsep}{8pt plus 0pt}%
  \setlength{\@dblfpbot}{0pt plus 1fil}%
}
\makeatother
\title{TARQ: Tail-Aware Reconstruction Quantization for Rare-Word Robust Automatic Speech Recognition}

\newcounter{noteYHctr} 

\newcounter{noteZiyu} 

\author{
  \textbf{Xinyu Wang\textsuperscript{1,2,*}},
  \textbf{Ziyu Zhao\textsuperscript{1,*}},
  \textbf{Ke Bai\textsuperscript{2}},
  \textbf{Silin Meng\textsuperscript{2}},
\\
  \textbf{Dongming Shen\textsuperscript{2}},
  \textbf{Xiao-Wen Chang\textsuperscript{1}},
  \textbf{Yixuan He\textsuperscript{3}}
\\
\\
  \textsuperscript{1}McGill University,
  \textsuperscript{2}Boson AI,
  \textsuperscript{3}Arizona State University
\\
  \thanks{
    \textsuperscript{*}Equal contribution. Emails:
    \href{mailto:xinyu@boson.ai}{xinyu@boson.ai},
    \href{mailto:ziyu.zhao2@mail.mcgill.ca}{ziyu.zhao2@mail.mcgill.ca}.
  }
}
\input{macros}
\begin{document}

\maketitle

\input{sections/00_abstract}

\section{Introduction}\label{sec:intro}
\input{sections/01_intro}

\section{Related Work}\label{sec:related}
\input{sections/02_related}

\section{Diagnosis: Frequency Inheritance in Data-Aware PTQ}\label{sec:diagnose}
\input{sections/03_diagnose}

\section{Tail-Aware Reconstruction Quantization}\label{sec:method}
\input{sections/04_method}

\section{Experiments}
\label{sec:experiments}
\input{sections/05_experiments}

\section{Discussion}\label{sec:disc}
\input{sections/06_discussion}

\section{Conclusion}\label{sec:conclusion}
\input{sections/07_conclusion}

\clearpage
\section*{Limitations}
\label{sec:limitations}
\input{sections/08_limitations}

\clearpage
\bibliography{custom}

\appendix
\clearpage
\newpage

\input{sections/A_appendix}

\end{document}

%% file: macros.tex
\newcommand{\TARQ}{\textsc{TARQ}}

\newcommand{\GPTQ}{GPTQ}
\newcommand{\AWQ}{AWQ}

\newcommand{\OmniQuant}{OmniQuant}
\newcommand{\SpQR}{SpQR}

\newcommand{\HQQ}{HQQ}
\newcommand{\SmoothQuant}{SmoothQuant}
\newcommand{\Hwz}{H_0}

\newcommand{\E}{\mathbb{E}}
\newcommand{\R}{\mathbb{R}}
\newcommand{\tail}{\mathrm{tail}}
\newcommand{\common}{\mathrm{common}}
\newcommand{\fp}{\mathrm{fp}}

\newcommand{\Wq}{\hat{W}}
\newcommand{\rareBAL}{\textsc{rareBAL}}

\newcommand{\rWER}{rare-WER}

\newcommand{\Wfour}{W4}
\newcommand{\Wthree}{W3}
\newcommand{\Wtwo}{W2}

%% file: sections/00_abstract.tex
\begin{abstract}
Data-aware post-training quantization (PTQ) minimizes a per-token
reconstruction loss on a small calibration corpus, implicitly
weighting positions by their empirical frequency. For
\textbf{A}utomatic \textbf{S}peech \textbf{R}ecognition (ASR), this
misaligns with tail-sensitive risk: names, numerals, and
domain-specific words receive proportionally little calibration mass.
We propose \textbf{Tail-Aware Reconstruction Quantization} (\TARQ), a
label-free PTQ framework that shifts calibration toward the lexical
tail via \textbf{\rareBAL}, a closed-form per-Linear-layer rule
equalizing common/tail mass, paired with a metric-consistent residual
correction. \TARQ\ requires no entity labels, no curated calibration
set, no validation decoding, and no additional training. Across eight
ASR backbones and six datasets at W4G128, \TARQ\ improves mean
rare-\textbf{W}ord \textbf{E}rror \textbf{R}ate (rare-WER) without an
aggregate-WER regression, achieves the lowest cross-corpus rare-WER
swing among compared methods, and transfers to entity-rich benchmarks
(ProfASR, ContextASR-Speech-En) without entity supervision.
\end{abstract}

%% file: sections/01_intro.tex
Large-scale end-to-end automatic speech recognition (ASR) models
based on Transformer architectures~\cite{vaswani2017attention,whisper,Qwen3-ASR,liu2025voxtral}
have substantially improved recognition accuracy, but their growing
size makes deployment increasingly constrained by memory bandwidth and
device capacity. Weight-only post-training quantization (PTQ) is an
attractive compression strategy: each Linear-layer full-precision
weight matrix $W$ is replaced by a low-bit (e.g., 4-bit integer)
approximation $\widehat W$ while activations stay in floating point,
and PTQ chooses $\widehat W$ from a small calibration set to keep the
per-layer output $\widehat W x$ close to $Wx$ without any gradient
updates or training-time data.

Most modern low-bit PTQ methods~\cite{gholami2022survey} were developed
primarily for large language models. Their data-aware variants share a
common principle: minimize a reconstruction loss averaged over
calibration token positions. This averaging encodes an implicit
calibration choice---because every position contributes equally to the
average, the effective calibration metric inherits the empirical
token-frequency distribution of the corpus, so frequent lexical patterns
dominate while infrequent ones contribute only in proportion to how
often they appear. This is appropriate when the deployment target is
average-token preservation, but it is a design choice rather than a
fundamental requirement, and it has not been examined for ASR.

This choice can be misaligned with tail-sensitive recognition risk.
Names, numerals, technical terms, and domain-specific vocabulary are a
recurring focus of ASR adaptation methods such as contextual
biasing~\cite{le21_interspeech,Tree-Constrained,sun2022minimising}, and
these positions tend to be lexically rare. Rare positions are plausible
candidates for fragility under quantization perturbation: they often
correspond to lower-frequency lexical decisions where a small weight
change may be more likely to alter the decoded token. Under a
frequency-weighted calibration metric, this fragile lexical-tail slice
receives only frequency-proportional optimization mass, and aggregate
WER further obscures any resulting degradation because its denominator
is dominated by common words. An ideal weighting would be derived from
per-position loss sensitivity (e.g., gradient or Fisher information),
but this requires labels and backward passes incompatible with the
one-pass nature of PTQ calibration.

We propose \textbf{Tail-Aware Reconstruction Quantization} (\TARQ), a
reconstruction-based PTQ framework that reweights the calibration metric
itself, using lexical rarity as a label-free proxy for fragile
positions. \TARQ\ has two closed-form components. \textbf{\rareBAL}
replaces the host PTQ calibration metric with a common/tail-balanced
one, where the balancing coefficient is set per Linear layer (i.e.,
each $y=Wx$ projection inside the model) by trace equalization. A scalar residual correction then keeps the sequential PTQ
sweep aligned with the reweighted objective. Both components reuse the
activation second-moment statistics that data-aware PTQ already
accumulates, adding no extra calibration pass.



\paragraph{Main contributions.}
\begin{itemize}[topsep=2pt,itemsep=2pt,leftmargin=*]
\item We identify an implicit position-weighting choice in
reconstruction-based PTQ: calibration metrics inherit the
corpus token-frequency distribution, misaligning with tail-sensitive
ASR risk in a way that aggregate WER hides.
\item We introduce \textbf{\TARQ}, a PTQ framework with two
closed-form components that reuse the second-moment statistics from a
single calibration pass: \textbf{\rareBAL}, a label-free
per-Linear-layer trace-equalization reweighting, and a metric-consistent scalar residual
correction.
\item Across eight ASR backbones, six datasets, and six calibration
corpora, \TARQ\ improves rare-WER on the majority of backbones,
remains competitive on plain WER, and is the most calibration-stable
method; entity-rich evaluations confirm that the rarity-based shift
transfers without entity supervision.
\end{itemize}

%% file: sections/02_related.tex
\paragraph{Weight-only PTQ for LLMs.}
Modern  PTQ has been developed primarily for large language
models. \GPTQ~\cite{gptq} minimizes a
per-column reconstruction loss under an inverse-Hessian metric;
\AWQ~\cite{awq} minimizes a per-channel salience-weighted reconstruction;
\OmniQuant~\cite{omniquant} jointly fits per-channel scales and clipping
bounds; \SpQR~\cite{dettmers2023spqr} keep a small fraction of weights at higher precision along the weight axis.
Despite differences in parameterization
and solver, all data-aware families share the same per-token activation
reconstruction loss, and are evaluated primarily on
LLM perplexity rather than ASR recognition risk. \TARQ\ addresses this objective-level mismatch by rebalancing the
calibration loss itself, so that rounding is no longer driven primarily
by high-frequency lexical mass.
\vspace{-5pt}
\paragraph{PTQ for ASR.}
PTQ for ASR remains less explored than LLM PTQ, and existing methods do
not resolve low-bit weight-only quantization for modern encoder-decoder
ASR backbones. Integer-only zero-shot quantization targets INT$8$
CTC/RNN-era models such as QuartzNet, Jasper, and
Conformer~\cite{kim2022integeronlyzeroshotquantizationefficient}.
StableQuant~\cite{hong2025stablequantlayeradaptiveposttraining} focuses
on CNN front-ends of encoder-only HuBERT/wav2vec~2.0 models and mainly
reports W$6$A$6$ results. GenPTQ~\cite{kang-kim-2025-genptq} studies
mixed-precision PTQ for Whisper and Conformer, but bit allocation acts
above the rounding objective and does not correct token-level imbalance.
Edge-ASR~\cite{liu2025edgeasr} benchmarks PTQ hosts on Whisper and
shows severe low-bit WER degradation, but leaves its cause and remedy
open. In contrast, \TARQ\ targets the shared calibration failure itself:
uniform reconstruction is frequency-aware but entity-agnostic. By
rebalancing common/tail optimization mass and solving the corrected
rounding problem, \TARQ\ directly optimizes for the rare-entity errors
that prior ASR PTQ methods do not address.

%% file: sections/03_diagnose.tex
\vspace{-0.5em}
Data-aware PTQ averages a per-position reconstruction loss across the
calibration corpus. This averaging encodes an implicit position
weighting: the calibration metric inherits the empirical token
frequency of the corpus, so lexical-tail positions enter the objective
only in proportion to how often they appear. We make this implicit
weighting explicit in two steps. \Cref{sec:diag-frequency} decomposes
the reconstruction loss into a frequency-weighted common/tail mixture
and shows its downstream signature: rare positions incur larger WER
degradation than common positions across PTQ solvers.
\Cref{sec:diag-mse-decouple} shows that optimizing the same
frequency-weighted objective more aggressively does not close this gap,
locating the common/tail asymmetry in the calibration metric itself,
not in how tightly that metric is minimized.
\begin{figure*}[!htbp]
\centering
\includegraphics[width=\linewidth]{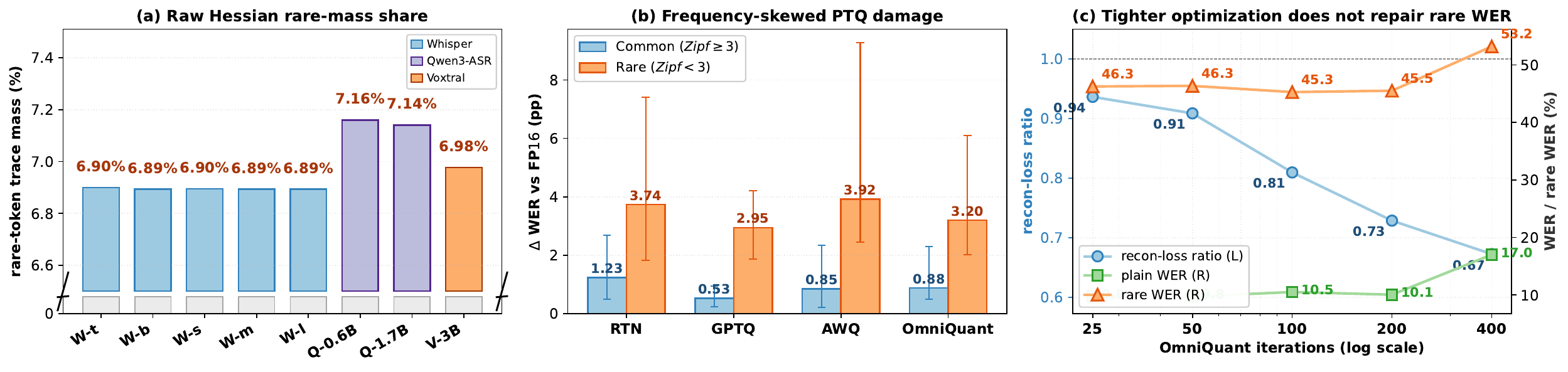}
\caption{\textbf{Diagnosis of frequency-weighted PTQ.}
(a)~\textbf{Raw Hessian rare-mass share}: across all eight backbones,
rare-token positions hold only a small fraction of the per-Linear-layer
calibration trace mass under the raw second-moment metric $H_\ell$
(median across decoder Linear layers).
(b)~\textbf{Frequency-skewed damage}: across PTQ solvers, rare
positions degrade more than common ones, the recognition-level
footprint of \cref{eq:recon-mixture}. Error bars: bootstrap $95\%$ CI.
(c)~\textbf{Tighter optimization does not repair rare-WER}:
\OmniQuant\ iteration sweep on Whisper-base drops the per-block
reconstruction-loss ratio substantially, but plain WER saturates and
\rWER\ can regress.}
\label{fig:diagnosis-row}
\end{figure*}
\subsection{Preliminaries}
\label{sec:diag-defs}

\paragraph{Weight-only uniform quantization.}
Weight-only PTQ replaces a Linear-layer weight $W\in\R^{m\times n}$
with a low-bit dequantized approximation $\Wq$. We use the standard
min-max symmetric scheme at bit-width $b$ and group size $g$: for each
row-wise group of $g$ contiguous input channels,
\begin{equation}
\begin{aligned}
\Wq &= s\cdot\mathrm{clamp}\!\left(\left\lfloor W/s\right\rceil,\;
-2^{b-1},\;2^{b-1}-1\right), \\
s &= \frac{\max W - \min W}{2^b-1},
\end{aligned}
\label{eq:quant}
\end{equation}
with $\lfloor\cdot\rceil$ round-to-nearest. The dequantized weight
$\Wq$ replaces $W$ at inference time. Lowering $b$ shrinks the model
footprint but coarsens the representable lattice, growing the
per-Linear-layer quantization error $\Delta W = W - \Wq$ and, in
general, increasing recognition error. The main experiments use
$b=4$, $g=128$ (W4G128).

\paragraph{Reconstruction loss.}
Let $\ell$ index the Linear layers of the model and let $N$ denote the
total number of calibration token positions; $x_{\ell,t}\in\R^n$ is
the activation entering layer $\ell$ at calibration position $t$.
Data-aware PTQ then chooses $\Wq$ to minimize the calibration
reconstruction loss
\begin{equation}
\mathcal{L}_{\mathrm{rec}}(\Delta W)
=
\frac{1}{N}\sum_{t=1}^{N}
\|\Delta W\, x_{\ell,t}\|_2^2.
\label{eq:recon-loss}
\end{equation}
The per-Linear-layer second moment used as the rounding metric in
second-order PTQ is $H_\ell = \sum_t x_{\ell,t} x_{\ell,t}^\top$;
normalization constants are omitted as they do not affect the rounding
geometry.

\paragraph{Lexical groups.}
A reference position is \textbf{rare} if its \texttt{wordfreq}~\cite{word_freq}
Zipf score (log-frequency per billion words) is below $3$, i.e., under
one occurrence per million words, and \textbf{common} otherwise. Let
$\mathcal{T},\mathcal{C}\subseteq\{1,\dots,N\}$ be the tail and common
position sets, with counts $N_\tail=|\mathcal{T}|$,
$N_\common=|\mathcal{C}|$, and let $g\in\{\common,\tail\}$ index the
group. The tail slice often includes named entities, numerals,
technical terms, and domain vocabulary, but rarity is a
lexical-frequency proxy rather than an entity label. We report plain
WER and rare-WER, restricting the reference denominator to tail
positions. Partitioning $H_\ell$ by group gives
$H_\ell = H^{\common}_\ell + H^{\tail}_\ell$, where
$H^{g}_\ell = \sum_{t \in g} x_{\ell,t} x_{\ell,t}^\top$.

\subsection{Frequency-weighted reconstruction mixture}
\label{sec:diag-frequency}

Let $p = N_\tail/N$ be the tail share. The group-wise reconstruction
loss for $g\in\{\common,\tail\}$ is
\begin{equation}
\mathcal{L}^{g}_\ell(\Delta W)
=
\frac{1}{N_g} \sum_{t\in g} \|\Delta W\, x_{\ell,t}\|_2^2 .
\label{eq:group-recon-loss}
\end{equation}
Then \cref{eq:recon-loss} decomposes as
\begin{equation}
\mathcal{L}_{\mathrm{rec}}
=
(1-p)\,\mathcal{L}^{\common}_\ell
+
p\,\mathcal{L}^{\tail}_\ell ,
\label{eq:recon-mixture}
\end{equation}
so each group enters the objective in proportion to its corpus
frequency. For the calibration sets used in our experiments, $p$ is
about $0.07$, so reducing $\mathcal{L}_{\mathrm{rec}}$ is dominated by
the common term, while $\mathcal{L}^{\tail}_\ell$ contributes
proportionally little. \Cref{eq:recon-mixture} thus exposes the
implicit position weighting quantitatively: it is fixed by corpus
statistics, and standard PTQ hyperparameters such as damping,
grouping, or iteration budget do not directly control the common/tail
balance.

The recognition-level signature matches this decomposition.
\Cref{fig:diagnosis-row}(b) splits 4-bit weight-only (W4) WER degradation relative to
the 16-bit \textbf{F}loating-\textbf{P}oint baseline (FP16) by reference-word frequency group. Across PTQ methods, rare-word
degradation is substantially larger than common-word degradation, even
though the methods use different rounding rules and calibration
heuristics. The group assigned small calibration mass exhibits
disproportionately large recognition degradation. Because the pattern
appears across solvers, it tracks the shared calibration mixture in
\cref{eq:recon-mixture} rather than any solver-specific artifact.
Aggregate (plain) WER masks this asymmetry because its denominator is
pooled over all reference tokens, which are common-dominated.

\subsection{Metric- vs solver-level asymmetry}
\label{sec:diag-mse-decouple}

A natural alternative explanation is that existing PTQ solvers simply
do not optimize the reconstruction objective tightly enough, and that
tighter optimization would close the common/tail gap. We test this by
sweeping the \OmniQuant\ iteration budget on Whisper-base. As shown in
\cref{fig:diagnosis-row}(c), increasing the iteration budget
substantially lowers the per-block reconstruction-loss ratio, from
$0.94$ to $0.67$. Downstream recognition quality does not track this
improvement: plain WER saturates, while \rWER\ does not improve and
can regress. Optimizing the same frequency-weighted objective more
tightly therefore does not shift the common/tail balance---the
asymmetry sits in the calibration metric, not in how tightly that
metric is minimized.

Tail-aware calibration must therefore intervene at the metric. Both
group terms $\mathcal{L}^{\common}_\ell$ and $\mathcal{L}^{\tail}_\ell$
are group averages (\cref{eq:group-recon-loss}); the metric below is
defined up to an irrelevant positive scale, which does not affect the
rounding optimum. We replace the frequency mixture in
\cref{eq:recon-mixture} with a group-rebalanced metric
\begin{equation}
\mathcal{L}^{\mathrm{bal}}_\ell(\Delta W)
=
\mathcal{L}^{\common}_\ell(\Delta W)
+
\lambda_\ell\,\mathcal{L}^{\tail}_\ell(\Delta W) ,
\label{eq:diag-loss-preview}
\end{equation}
where $\lambda_\ell$ is a per-Linear-layer coefficient constructed in
\cref{sec:method-rarity}. \Cref{eq:diag-loss-preview} previews the
metric-level shift; \cref{sec:method-rarity} gives the closed-form
trace equalization used to set $\lambda_\ell$.

%% file: sections/04_method.tex
The diagnosis in \cref{sec:diagnose} points to a metric-level shift:
replace the frequency-weighted calibration metric with a
common/tail-balanced one. We instantiate this in \textbf{Tail-Aware Reconstruction
Quantization} (\TARQ), a reconstruction-based PTQ framework centered on
one shared quadratic form. The first component is \textbf{\rareBAL},
which equalizes rare and common trace mass via one scalar per Linear layer
(\cref{sec:method-rarity}). The second component is a propagation-aware
residual correction that builds a target for the sequential PTQ solver
under the same rebalanced metric (\cref{sec:method-solver}). The two
are not independent: \rareBAL\ sets the per-Linear-layer objective, while the
residual step keeps the sequential layer sweep metric-consistent after
earlier layers have been quantized.

For each Linear layer $\ell$, \TARQ\ solves an in-lattice projection
problem under the \rareBAL\ metric:
\begin{equation}
\widehat W_\ell
=
\arg\min_{\widetilde W\in\mathcal Q}
\left\|
\left(
W_\ell + \alpha_\ell D_\ell - \widetilde W
\right)
\left(H^{\mathrm{rB}}_\ell\right)^{1/2}
\right\|_F^2 ,
\label{eq:tarq-objective}
\end{equation}
where $\mathcal Q$ is the 4-bit weight, group-size-128 (W4G128)
lattice, $\alpha_\ell$ is a per-Linear-layer scalar
(\cref{eq:alpha}), $D_\ell$ is the propagation-aware direction
(\cref{eq:direction-target}), and $H^{\mathrm{rB}}_\ell$ is the
\rareBAL\ metric. The residual target $W_\ell + \alpha_\ell D_\ell$
is then projected back into the W4G128 lattice, so the returned
weight remains in $\mathcal Q$ (no floating-point residual stored).

\subsection{\rareBAL: rare-balanced calibration metric}
\label{sec:method-rarity}

The implicit position weighting exposed in \cref{eq:recon-mixture}
belongs to a general family of frequency-bucketed metrics
$H_\ell(\alpha) = \sum_f \alpha_f \bar H_\ell^f$, where $\bar H_\ell^f$
is the (trace-normalized) second moment of positions in frequency
bucket $f$ and $\alpha_f\geq 0$ are bucket weights. Standard PTQ
corresponds to $\alpha_f$ proportional to bucket size; the
loss-sensitivity-aware ideal would set $\alpha_f$ from per-bucket
gradient or Fisher information. \rareBAL\ instantiates the simplest
nontrivial instance in this family: a two-bucket partition (rare,
common) with weights set by closed-form trace equalization, requiring
neither labels nor a held-out signal.

Using the notation of \cref{sec:diag-defs}, with $x_t$ the activation
entering Linear layer $\ell$ in the partially quantized network, the
per-group second-moment matrices are
\begin{equation}
H^{\common}_\ell
=
\sum_{t\in\mathcal C} x_t x_t^\top,
\qquad
H^{\tail}_\ell
=
\sum_{t\in\mathcal T} x_t x_t^\top .
\label{eq:group-h}
\end{equation}
Their traces measure the total calibration mass contributed by each
group. \rareBAL\ rescales the rare slice by a single scalar:
\begin{equation}
\lambda_\ell
=
\frac{\operatorname{tr}(H^{\common}_\ell)}
     {\operatorname{tr}(H^{\tail}_\ell)+\varepsilon},
H^{\mathrm{rB}}_\ell
=
H^{\common}_\ell
+
\lambda_\ell H^{\tail}_\ell .
\label{eq:rarebal}
\end{equation}
By construction,
\begin{equation}
\operatorname{tr}\!\left(\lambda_\ell H^{\tail}_\ell\right)
=
\operatorname{tr}\!\left(H^{\common}_\ell\right).
\end{equation}
The choice of trace is tied to the reconstruction objective: for group
$g\in\{\common,\tail\}$, the group contribution to the weighted
reconstruction loss is
\begin{equation}
\mathcal L^g_\ell(\Delta W)
=
\operatorname{tr}
\left(
\Delta W\, H^g_\ell\, \Delta W^\top
\right),
\Delta W = W-\widehat W .
\label{eq:group-loss}
\end{equation}
Thus the trace indexes the total second-moment mass that each group
contributes to the quadratic metric. \Cref{eq:rarebal} is the unique
single-scalar rescaling that equalizes rare and common trace mass. The
resulting 50/50 split is algebraic rather than a normative claim about
deployment cost; an alternative cost ratio $c$ can be obtained by using
$c\lambda_\ell$ instead of $\lambda_\ell$. we sweep $c\in[0.25, 4]$
in \cref{sec:component-analysis} and find that the trace-equalization
default $c=1$ is empirically near-optimal across backbones, and
markedly the best choice on Qwen3-ASR-0.6B. Because both traces shift
with the calibration corpus, $\lambda_\ell$ adapts automatically, which
helps explain the cross-corpus stability reported in
\cref{sec:exp-cross-calib}.

\subsection{Propagation-aware residual correction}
\label{sec:method-solver}

\rareBAL\ specifies the metric for a single Linear layer, but a PTQ solver
visits layers sequentially: by the time layer $\ell$ is reached, the
input activation has drifted from the full-precision trajectory.
Standard GPTQ lets the next layer absorb this drift implicitly; once
\rareBAL\ changes the reconstruction geometry, the propagation error
should be corrected in the same geometry.

Let $x_t^{\fp}$ and $x_t$ denote the full-precision and
partially-quantized activations at layer $\ell$, respectively. The input-error
cross-moment $H_{\Delta,\ell} = \sum_{t}(x_t^{\fp}-x_t)x_t^\top$
defines a propagation direction and continuous target under
$H^{\mathrm{rB}}_\ell$:
\begin{equation}
D_\ell = W_\ell H_{\Delta,\ell}
\left(H^{\mathrm{rB}}_\ell+\delta I\right)^{-1},
W^{\mathrm{tar}}_\ell = W_\ell+\alpha_\ell D_\ell .
\label{eq:direction-target}
\end{equation}
The final weight is the in-lattice projection
$\widehat W_\ell = \Pi^{H^{\mathrm{rB}}_\ell}_{\mathcal Q}
(W^{\mathrm{tar}}_\ell)$, where $\Pi^{H}_{\mathcal Q}$ is a
GPTQ-style projection under metric $H$. The scalar $\alpha_\ell$ is
fit by one-dimensional least squares against a pilot displacement
$E_\ell = \Pi^{H^{\mathrm{rB}}_\ell}_{\mathcal Q}(W_\ell) - W_\ell$:
\begin{equation}
\alpha_\ell^\star
=
\frac{\langle E_\ell,D_\ell\rangle_{H^{\mathrm{rB}}_\ell}}
     {\langle D_\ell,D_\ell\rangle_{H^{\mathrm{rB}}_\ell}+\varepsilon},
\quad
\langle A,B\rangle_H = \operatorname{tr}(AHB^\top).
\label{eq:alpha}
\end{equation}

The residual step is not a second reweighting mechanism but a
metric-consistent solver refinement: the correction is computed
under the same $H^{\mathrm{rB}}_\ell$ so the sequential sweep does
not revert to the original frequency-weighted geometry. Consistent
with this solver-level role, the residual contributes a
backbone-dependent refinement on top of \rareBAL\ rather than a
separate rarity-aware effect (\cref{sec:component-analysis}).

\subsection{Algorithm}
\label{sec:method-pipeline}

\Cref{alg:tarq} summarizes \TARQ\ for one Linear layer: \rareBAL\
builds the metric $H^{\mathrm{rB}}$, a pilot GPTQ pass produces a
first quantization $\widehat W^{(0)}$, and the residual step
constructs a propagation-aware target which is projected back into
the same W4G128 lattice under the same $H^{\mathrm{rB}}$ (no
floating-point residual stored). We write $\mathrm{GPTQ}(W,H,b,g)$
for a standard GPTQ sweep projecting $W$ onto the $(b,g)$ lattice
under metric $H$.

\begin{algorithm}[t]
\small
\DontPrintSemicolon
\SetKwInOut{KwIn}{Input}
\SetKwInOut{KwOut}{Output}

\KwIn{$W\in\mathbb R^{m\times n}$; paired activations
$\{x_t^{\fp},x_t\}_{t=1}^{N}$ with tags $\mathcal T,\mathcal C$;
config $(b,g,\delta)$.}
\KwOut{$\widehat W\in\mathcal Q$.}

$H^{\common}\leftarrow\sum_{t\in\mathcal C}x_t x_t^\top,\;
H^{\tail}\leftarrow\sum_{t\in\mathcal T}x_t x_t^\top$\;
$H_{\Delta}\leftarrow
\sum_{t=1}^{N}(x_t^{\fp}-x_t)x_t^\top$\;
$\lambda\leftarrow\operatorname{tr}(H^{\common})/
(\operatorname{tr}(H^{\tail})+\varepsilon)$;\;
$H^{\mathrm{rB}}\leftarrow H^{\common}+\lambda H^{\tail}$\;
$\widehat W^{(0)}\leftarrow\mathrm{GPTQ}(W,H^{\mathrm{rB}},b,g)$\;
$G\leftarrow (H^{\mathrm{rB}}+\delta I)^{-1}$;\;
$D\leftarrow W H_{\Delta}G$;\;
$E\leftarrow \widehat W^{(0)}-W$\;
$\alpha^\star\leftarrow
\operatorname{tr}(E H^{\mathrm{rB}}D^\top)/
(\operatorname{tr}(D H^{\mathrm{rB}}D^\top)+\varepsilon)$\;
$W^{\mathrm{tar}}\leftarrow W+\alpha^\star D$\;
$\widehat W\leftarrow\mathrm{GPTQ}(W^{\mathrm{tar}},H^{\mathrm{rB}},b,g)$\;

\Return $\widehat W$\;
\caption{Per-Linear-layer \TARQ.}
\label{alg:tarq}
\end{algorithm}

%% file: sections/05_experiments.tex
\subsection{Setup}
\label{sec:exp-setup}

\textbf{Backbones.}
We evaluate eight ASR backbones spanning three architectures:
Whisper-tiny/base/small/medium/large~\cite{whisper},
Qwen3-ASR-$0.6$B/$1.7$B~\cite{Qwen3-ASR}, and Voxtral-Mini-$3$B-$2507$~\cite{liu2025voxtral}.

\textbf{Quantization.}
All methods are evaluated under weight-only \Wfour\ G$128$
quantization. We compare against \GPTQ~\cite{gptq}, \AWQ~\cite{awq}, and
\OmniQuant~\cite{omniquant}, when compatible.

\textbf{Calibration.}
Unless otherwise stated, calibration uses $128$ utterances. We test
six calibration corpora: LibriSpeech-train.100 (LS-clean), SPGI-S
train, VoxPopuli-en train, and three rare-biased calibrations
(\texttt{r-top}, \texttt{r-mix}, \texttt{r-cross}) constructed by
sampling utterances on their \texttt{wordfreq} Zipf-rare token density;
see \cref{app:cross} for the construction.

\textbf{Evaluation.}
We evaluate on six datasets per backbone: LibriSpeech-clean~\cite{panayotov2015librispeech},
LibriSpeech-other~\cite{panayotov2015librispeech}, SPGI~\cite{oneil2021spgispeech}, VoxPopuli~\cite{wang-etal-2021-voxpopuli}, GigaSpeech~\cite{chen2021gigaspeech}, and TED-LIUM~\cite{hernandez2018ted}.
The main metrics are plain WER and \rWER, where \rWER\
restricts the reference denominator to words with \texttt{wordfreq}~\cite{word_freq}
Zipf $<3$. Per-cell results are reported in \cref{app:full-tables}.

\begin{figure*}[!t]
\centering
\includegraphics[width=\textwidth]{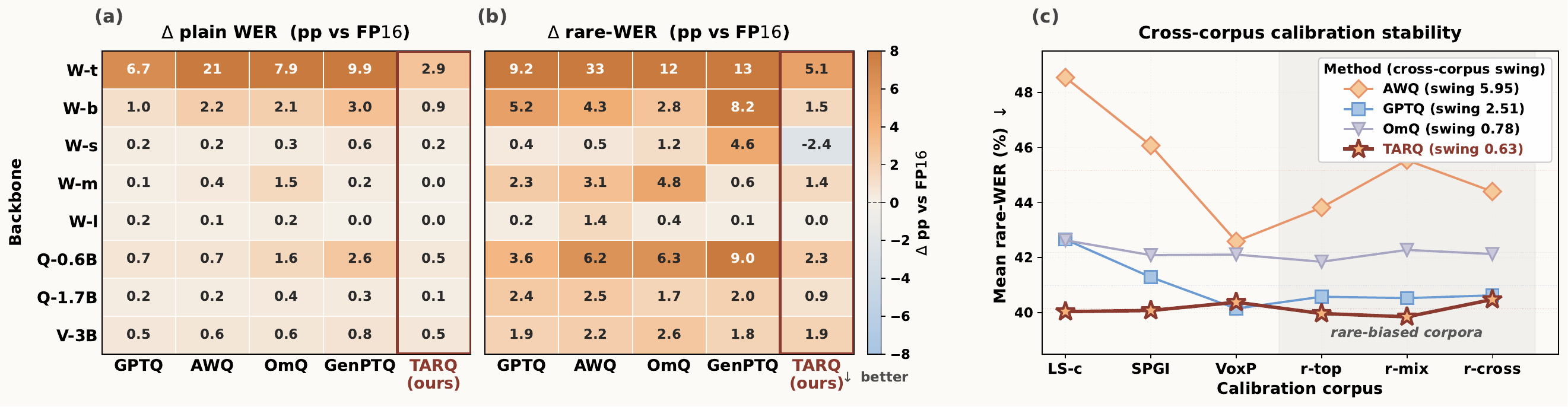}
\caption{\textbf{(a, b) Headline accuracy across 8 backbones $\times$
5 PTQ methods} (W4G128, mean over 6 ASR test sets). Cells show
$\Delta$ vs FP$16$; the \TARQ\ column (boxed) is consistently closest
to FP$16$ on the rare-WER side, while the plain-WER side is uniformly
pale across methods---the asymmetry sits in rare-WER.
\textbf{(c) Cross-corpus calibration stability}: rare-WER as a
function of calibration corpus; legend gives cross-corpus swing
($\max{-}\min$). \TARQ\ collapses to $0.63$ percentage points (pp) swing,
vs $2.51$ / $5.95$ / $0.78$ for \GPTQ\,/\,\AWQ\,/\,\OmniQuant.}
\label{fig:main-and-cross}
\end{figure*}

\subsection{Main result}
\label{sec:exp-main}

\Cref{fig:main-and-cross}(a,b) visualizes per-backbone $\Delta$WER
under LS-clean calibration. Per-backbone means are reported in
\cref{tab:lsclean-compact} and the full per-dataset breakdown in
\cref{app:full-tables}. Two patterns are
immediately visible. The right (rare-WER) panel is uniformly more
saturated than the left across all baselines---the visual signature
of the frequency-weighted mismatch diagnosed in
\cref{sec:diagnose}, hidden by aggregate WER but unmasked once the
lexical tail is scored separately. The \TARQ\ column, by contrast, is
the lightest column in both panels on nearly every row: \TARQ\
attains the best mean plain WER on all eight backbones and the best
mean rare-WER on six, with the remaining two cells still beating
every W4G128 weight-only baseline. Gains are the largest on the
Whisper-tiny/-base backbones, where the uniform calibration metric
most visibly underweights rare tokens, and shrink gracefully on
larger models where the FP16-to-quantized gap is already narrow for
all methods. Critically, plain WER tracks rare-WER on every
backbone---\TARQ\ closes the lexical-tail gap without a plain-WER
trade-off.

\begin{table}[t]
\centering
\caption{\textbf{Mean WER across six datasets under LS-clean
calibration} (LS-c, LS-o, SPGI, VoxP, Giga, TedL). Each cell reports
plain WER (\%) / rare-WER (\%). All quantized methods use W4G128
weight-only, except GenPTQ (mixed-precision, 4-bit average). Bold:
rank-1 among quantized methods per (backbone, metric). FP16 reference
and full per-dataset breakdown in \cref{tab:lsclean-full}.}
\label{tab:lsclean-compact}
\setlength{\tabcolsep}{3pt}
\renewcommand{\arraystretch}{1.05}
\footnotesize
\begin{tabular}{l rrrr >{\columncolor{gray!15}}r}
\toprule
& \multicolumn{1}{c}{GPTQ} & \multicolumn{1}{c}{AWQ} & \multicolumn{1}{c}{OmQ} & \multicolumn{1}{c}{GenPTQ} & \multicolumn{1}{c}{\textbf{\TARQ}} \\
\midrule
\multicolumn{6}{l}{\textit{Plain WER (\%)}} \\
W-t    & 17.86 & 31.86 & 19.07 & 21.08          & \textbf{14.03} \\
W-b    & 9.48  & 10.64 & 10.52 & 11.45          & \textbf{9.31}  \\
W-s    & 7.22  & 7.27  & 7.33  & 7.59           & \textbf{7.19}  \\
W-m    & 6.52  & 6.83  & 7.95  & 6.69           & \textbf{6.49}  \\
W-l    & 6.07  & 6.04  & 6.12  & \textbf{5.95}  & \textbf{5.95}  \\
Q-0.6B & 6.76  & 6.79  & 7.64  & 8.71           & \textbf{6.57}  \\
Q-1.7B & 5.93  & 5.91  & 6.11  & 6.05           & \textbf{5.84}  \\
V-3B   & 7.16  & 7.23  & 7.19  & 7.38           & \textbf{7.14}  \\
\midrule
\multicolumn{6}{l}{\textit{Rare-WER (\%)}} \\
W-t    & 60.04 & 84.10 & 62.59 & 63.82          & \textbf{55.93} \\
W-b    & 47.57 & 46.75 & 45.27 & 50.64          & \textbf{43.96} \\
W-s    & 38.87 & 38.95 & 39.65 & 43.04          & \textbf{36.03} \\
W-m    & 35.21 & 36.00 & 37.70 & \textbf{33.49} & 34.28          \\
W-l    & 29.23 & 30.47 & 29.39 & 29.18          & \textbf{29.07} \\
Q-0.6B & 34.11 & 36.66 & 36.73 & 39.50          & \textbf{32.78} \\
Q-1.7B & 26.57 & 26.65 & 25.94 & 26.19          & \textbf{25.10} \\
V-3B   & 27.09 & 27.33 & 27.75 & \textbf{26.96} & 27.04          \\
\bottomrule
\end{tabular}
\end{table}

\subsection{Cross-calibration robustness}
\label{sec:exp-cross-calib}

\Cref{fig:main-and-cross}(c) sweeps six calibration corpora and
reports the four-backbone six-dataset mean rare-WER, alongside the
cross-corpus swing ($\max{-}\min$). \TARQ\ is both the lowest and the
flattest line in the figure---rank-1 on $5/6$ corpora and with the
smallest swing among compared methods. \OmniQuant\ reaches its
stability by learning scale transformations; \TARQ\ delivers tighter
stability from a single closed-form trace ratio read off the same
activations the solver already processes, with no auxiliary
optimization. Manually enriching calibration with rare words helps
\GPTQ, but \TARQ\ on ordinary LS-clean already sits below \GPTQ's best
rare-WER across any corpus---the curation gain is real but small, and
\TARQ\ captures it without curation. The per-corpus breakdown is
reported in \cref{app:cross}.

\subsection{Transfer to entity-rich benchmarks}
\label{sec:exp-entity-bench}

Zipf rarity is a label-free proxy. To check whether the calibration
correction transfers to explicitly entity-rich data, we evaluate \TARQ\ on
ProfASR (professional-domain speech)~\cite{piskala2025profasrbench} and ContextASR-Speech-En
(contextual-entity-enriched)~\cite{wang2025asrbench}. Calibration is LS-clean throughout; no
entity labels are used.

\begin{table}[!htbp]
\centering
\caption{\textbf{Transfer to entity-rich benchmarks} under LS-clean
calibration. \textbf{P} = plain WER (\%); \textbf{R} = rare-WER (\%).
Bold: rank-1 among quantized; underline: rank-2; FP16 in italics.}
\label{tab:entity-bench}
\setlength{\tabcolsep}{3pt}
\renewcommand{\arraystretch}{0.95}
\scriptsize
\begin{tabular}{l rr rr rr rr}
\toprule
& \multicolumn{4}{c}{W-base} & \multicolumn{4}{c}{Q3-1.7B} \\
\cmidrule(lr){2-5}\cmidrule(lr){6-9}
& \multicolumn{2}{c}{ProfASR} & \multicolumn{2}{c}{ContextASR}
& \multicolumn{2}{c}{ProfASR} & \multicolumn{2}{c}{ContextASR} \\
\cmidrule(lr){2-3}\cmidrule(lr){4-5}\cmidrule(lr){6-7}\cmidrule(lr){8-9}
method & P & R & P & R & P & R & P & R \\
\midrule
\textit{FP16} & \textit{13.22} & \textit{52.38} & \textit{45.74} & \textit{67.54} & \textit{6.13} & \textit{17.01} & \textit{7.26} & \textit{13.57} \\
\midrule
RTN       & 13.86 & 56.35 & 50.48 & 72.41 & 7.37 & 22.95 & 10.62 & 19.69 \\
GPTQ      & \underline{13.41} & \underline{54.97} & \underline{46.42} & \underline{69.60} & \textbf{6.45} & 19.65 & \underline{8.89} & \underline{17.67} \\
AWQ       & 14.94 & 57.48 & 49.02 & 72.36 & 6.53 & \underline{19.44} & 11.33 & \underline{17.67} \\
OmniQuant & 13.95 & 55.45 & 51.23 & 72.92 & 6.83 & 19.94 & 10.24 & 18.59 \\
\midrule
\rowcolor{gray!15}
\textbf{\TARQ\ } & \textbf{12.92} & \textbf{54.18} & \textbf{45.32} & \textbf{68.70} & \textbf{6.45} & \textbf{18.96} & \textbf{8.77} & \textbf{17.56} \\
\bottomrule
\end{tabular}
\end{table}

\TARQ\ takes rank-1 on every (backbone, dataset, metric) cell, winning
or tying the best quantized baseline on all four plain-WER cells and
all four rare-WER cells. On W-base, gains over \GPTQ\ are
$0.49$/$1.10$~pp plain and $0.79$/$0.90$~pp rare
(ProfASR/ContextASR); on Q-1.7B, \TARQ\ ties \GPTQ\ on ProfASR plain
($6.45$), leads ContextASR plain by $0.12$~pp, and improves rare-WER
by $0.69$ and $0.11$~pp. The rarity-based correction transfers to
entity-rich evaluation without entity supervision.

\paragraph{Qualitative effect on rare-word recovery.}
\Cref{fig:hero-examples} shows two utterance-level cases where
\TARQ\ recovers a rare reference word that every other W4G128
baseline drops; \cref{app:qualitative} reports more across all
three ablated backbones.

\begin{figure}[!htbp]
\centering
\begin{rwexamplebox}
\textbf{\texttt{marmalades}} \quad\textit{Whisper-base, LS-other}\\[2pt]
\rwRef \rwWord{marmalades} jams and fruit pastes are of the same nature \ldots\\
\rwOK~\rwTARQ: \rwWord{marmalades} jams and fruit paces of the same nature \ldots\\
\rwNO~\GPTQ: \emph{marmalade} jams and fruit \emph{paste} are the same nature \ldots\\
\rwNO~\AWQ: \emph{marmalade} jams and fruit \emph{pace} are the same nature \ldots\\
\rwNO~\OmniQuant: \emph{marmalade} and fruit \emph{paster}\\
\rwNO~GenPTQ: \emph{marmalade is jams and fruit paste} \ldots
\end{rwexamplebox}
\begin{rwexamplebox}
\textbf{\texttt{dandan}} \quad\textit{Qwen3-ASR-0.6B, LS-other}\\[2pt]
\rwRef \ldots the sultan commanded his wazir \rwWord{dandan} call a ten days halt \ldots\\
\rwOK~\rwTARQ: \ldots his wazir \rwWord{dandan} call a ten days halt \ldots\\
\rwNO~\GPTQ: \ldots his \emph{vizier done} call at ten days halt \ldots\\
\rwNO~\OmniQuant: \ldots his \emph{wazirdandan} call a ten days halt \ldots\\
\rwNO~GenPTQ: \ldots his \emph{wazirdan khan} call at ten days halt \ldots
\end{rwexamplebox}
\caption{\textbf{Rare-word recovery, two real utterances.} \TARQ\
preserves the rare reference word (\rwWord{bold orange}); other
W4G128 baselines substitute (\emph{italic}) or degenerate. Truncated
$\sim$$10$-word windows around the rare word.}
\label{fig:hero-examples}
\end{figure}

\subsection{Component analysis}
\label{sec:component-analysis}

\TARQ\ has two pieces: the \rareBAL\ metric rebalance
(\cref{sec:method-rarity}) and a scalar residual correction
(\cref{sec:method-solver}). \Cref{tab:ablation} applies each in
isolation. \rareBAL\ alone does most of the work, closing the
bulk of \TARQ's rare-WER gain on both backbones; the residual
correction provides a smaller, backbone-dependent refinement on
top. Combined gains are sub-additive, consistent with the two
components correcting overlapping aspects of the same calibration
imbalance---\rareBAL\ rebalances the rounding objective at each
Linear layer, while the residual step keeps the sequential sweep aligned
with that rebalanced metric.

\begin{table}[!htbp]
\centering
\caption{\textbf{Component ablation} on W-b and Q-0.6B under W4G128
with LS-clean calibration. Mean plain WER / rare-WER (\%) across six
datasets. \textit{+rareBAL}: rebalanced metric alone;
\textit{+residual}: residual correction alone; \TARQ: both. Bold: rank-1
per (backbone, metric); underline: rank-2. Per-dataset breakdown in
\cref{tab:ablation-full}.}
\label{tab:ablation}
\setlength{\tabcolsep}{6pt}
\renewcommand{\arraystretch}{1.0}
\small
\begin{tabular}{llrr}
\toprule
Model & Method & Plain WER & Rare-WER \\
\midrule
\multirow{4}{*}{W-b}
& GPTQ           & 9.48          & 47.58 \\
& +rareBAL       & \textbf{9.26} & \underline{44.63} \\
& +residual      & 9.38          & 45.07 \\
\rowcolor{gray!15}
& \textbf{\TARQ\ } & \underline{9.31} & \textbf{43.96} \\
\midrule
\multirow{4}{*}{Q-0.6B}
& GPTQ           & 6.76          & 34.11 \\
& +rareBAL       & \textbf{6.56} & \underline{32.80} \\
& +residual      & 6.62          & 33.29 \\
\rowcolor{gray!15}
& \textbf{\TARQ\ } & \underline{6.57} & \textbf{32.78} \\
\bottomrule
\end{tabular}
\end{table}

Two further controls---a reweighting-source ablation (rare vs random
vs common upweighting) and a cost-ratio sweep around the trace-
equalization default $c=1$---confirm that the rarity signal is what
matters and that the default is robust; both are deferred to
\cref{app:weighting-and-c-sweep}. Robustness to the Zipf threshold
and integration with \SmoothQuant~\cite{xiao2023smoothquant} and
\SpQR~\cite{dettmers2023spqr} are reported in
\cref{app:zipf_robustness,app:integration}.

\subsection{Deployment profile}
\label{sec:deployment}

\begin{figure}[t]
\centering
\includegraphics[width=\linewidth]{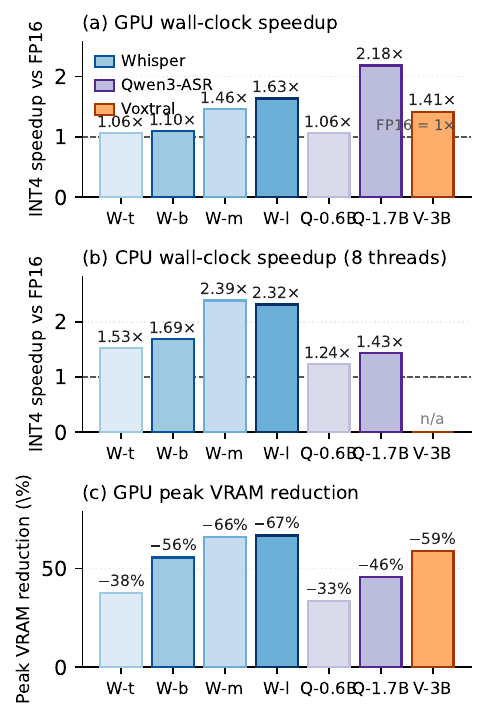}
\caption{\textbf{Deployment profile.} (a)~INT4 wall-clock speedup
over FP16 on a single A100; (b)~the same comparison on AMD EPYC
7V12 at $8$~threads (Voxtral has no CPU run); (c)~GPU peak VRAM
reduction. GPU wall-clock uses $30$\,s audio for W-large-v3 and
Voxtral (end-to-end), $11$\,s for the other Whisper sizes and
Qwen3-ASR; CPU rows are $11$\,s. Full per-backbone benchmarks (all
$11$\,s and $30$\,s rows, CPU latency at $\{4,8,16\}$ threads, and
max-RSS) in \cref{app:efficiency}.}
\label{fig:eff-headline}
\end{figure}

\TARQ\ produces standard W4G128 weight-only models, so deployment
gains are shared with any 4-bit scheme; \TARQ's contribution is
delivering them at the lower rare-WER reported above
(\cref{fig:eff-headline}). On a single A100, INT4 yields
$1.43$/$2.18\times$ end-to-end speedups on Voxtral-Mini-3B and
Qwen3-ASR-1.7B ($-59\%$ and $-46\%$ peak VRAM, respectively). On CPU
(\texttt{whisper.cpp}~\citep{whisper_cpp},
\texttt{qwen3-asr.cpp}~\citep{qwen3asrcpp}, AMD EPYC), INT4 tips
Whisper-large-v3 into real time (RTF $0.87$ at $8$ threads, vs
$2.01$ FP16) and halves max-RSS. Full per-backbone benchmarks in
\cref{app:efficiency}.

\FloatBarrier

%% file: sections/06_discussion.tex
\subsection{Applicability}
\label{sec:disc-regime}

\TARQ\ is effective when the calibration metric assigns small mass
to rare-token directions, so the uniform reconstruction objective
is dominated by common-token directions. Gains are the largest on
Whisper-style backbones; on the larger Qwen3-ASR and Voxtral
backbones the W$4$-vs-FP16 rare-WER gap under vanilla GPTQ is
already small, leaving less headroom for \rareBAL. The applicable
regime is determined not by parameter count alone but by whether
the calibration metric is rare-mass-starved. \rareBAL\ on ordinary
LS-clean also matches or beats vanilla GPTQ with manually
rare-enriched calibration (\cref{tab:cross-calib-summary}),
delivering the lexical-tail shift without a curated calibration pool.

Concretely, across the six calibration corpora in
\cref{tab:cross-calib-summary} \TARQ\ holds a cross-corpus rare-WER
swing of $0.63$~pp at a $40.13\%$ mean, against $0.78$/$42.18$ for
\OmniQuant, $2.51$/$40.97$ for \GPTQ, and $5.95$/$45.16$ for \AWQ:
the metric-level rebalance is a tighter knob than swapping the
calibration pool. Under LS-clean calibration
(\cref{tab:lsclean-full}), \TARQ\ achieves rank-$1$ mean plain~WER on
all eight backbones and rank-$1$ mean rare-WER on six of eight,
with the two remaining cells (Whisper-medium, Voxtral-3B) taken by
mixed-precision GenPTQ by a thin margin.

\subsection{Relation to sensitivity-aware calibration}
\label{sec:disc-sensitivity}

\rareBAL\ uses lexical rarity as a label-free proxy for fragile
positions; the more principled choice would be a sensitivity-aware
weighting derived from per-position loss gradients or Fisher
information, which subsumes \rareBAL\ as a special case of the
family $H_\ell(\alpha){=}\sum_f \alpha_f \bar H_\ell^f$
(\cref{sec:method-rarity}). We do not claim rarity is optimal; we
claim it is a cheap, label-free, single-pass proxy that captures
most of the achievable rare-WER gain in our setting. A direct
treatment of sensitivity-aware $\alpha_f$ is left to future work.

%% file: sections/07_conclusion.tex
We propose \textbf{Tail-Aware Reconstruction Quantization} (\TARQ):
\textbf{\rareBAL}, a closed-form per-Linear-layer rebalance that
equalizes the common/tail calibration mass, plus a metric-consistent
residual correction. Across eight ASR backbones, \TARQ\ improves
rare-WER on the majority of backbones while remaining competitive on
plain WER, posts the smallest cross-corpus rare-WER swing among
compared fixed-hyperparameter baselines, and transfers to entity-rich
benchmarks (ProfASR, ContextASR-Speech-En). Lower bit-widths
(\Wthree/\Wtwo) and direct loss-sensitivity weighting are natural
next steps.

%% file: sections/08_limitations.tex
\paragraph{Method scope.} \TARQ\ targets data-aware reconstruction
PTQ for ASR: it consumes a calibration-set second moment $H_\ell$
and rebalances its trace mass in closed form
(\cref{sec:method-rarity}). The method therefore does not apply
unchanged to data-free PTQ (e.g.,
\HQQ~\cite{badri2023hqq}), rotation/lattice quantization
(\texttt{QuIP\#}~\cite{tseng2024quipbetterllmquantization},
\texttt{QTIP}~\cite{tseng2024qtip}), or activation-quantized W$4$A$4$
settings, none of which expose a per-Linear-layer reconstruction
metric in the same form. Whether the common/tail asymmetry
documented in \cref{sec:diag-frequency} persists outside
reconstruction-style PTQ is open.

\paragraph{Bit-widths.} All experiments use W$4$G$128$. Lower
bit-widths (\Wthree/\Wtwo) likely amplify the common/tail mass
imbalance, but we did not measure whether the closed-form trace
equalization remains well-conditioned in those regimes, where the
representable scale grid is much coarser and rounding error is no
longer dominated by the dense interior of the weight distribution.

\paragraph{Rarity proxy.} \rareBAL\ uses lexical rarity
(\texttt{wordfreq} Zipf score below~$3$) as a label-free proxy for
fragile positions. A principled choice of $\alpha_f$ in
$H_\ell(\alpha){=}\sum_f \alpha_f \bar H_\ell^f$ would come from
per-position loss sensitivity (gradient or Fisher information), but
that requires labels and a backward pass. The rarity proxy is cheap
and single-pass, but it may under-weight fragile positions that
happen to be lexically common and over-weight rare positions that
are already easy; we treat it as a useful approximation, not as the
optimal choice.

\paragraph{Failure mode under contextless rare nouns.} On short
utterances dominated by a single rare proper noun, \TARQ's
layer-level rebalance can substitute the rare token with a
phonetically close high-frequency neighbour
(e.g., \texttt{bartley}\,$\to$\,\emph{partly}); the layer-level
rebalance has no mechanism to recover the surface form at decoding
time. \Cref{app:failure} gives a worked example and discusses why
patching this at the calibration metric would risk regressing the
common/tail balance. Inference-time contextual biasing is a
complementary mitigation that we did not combine with \TARQ\ in
this work.

\paragraph{Evaluation conventions.} Rare-WER is reported with
insertion errors assigned to the aligned reference position, so
insertions adjacent to rare references count as rare-relevant.
Alternative alignment conventions (e.g., charging insertions to the
nearest common neighbour) shift absolute rare-WER numbers but apply
uniformly across compared methods. The Zipf threshold of~$3$ is
the standard one-per-million-words boundary used by
\texttt{wordfreq}~\cite{word_freq}; \cref{app:zipf_robustness} reports the
robustness of the ordering to this threshold.

\paragraph{Languages and domains.} The main experiments are
English-only and use general-domain read-speech and parliamentary
ASR corpora. We did not evaluate on tonal languages, code-switched
audio, or vertical domains (medical, legal, technical) where rarity
statistics from \texttt{wordfreq} may not transfer well. We expect
the diagnosis in \cref{sec:diag-frequency} to remain qualitatively
valid wherever the calibration corpus is common-token-dominated,
but absolute rare-WER gains in those settings are not verified.

\paragraph{Comparator coverage.} We compare against
fixed-hyperparameter weight-only PTQ baselines (\GPTQ, \AWQ,
\OmniQuant) and the mixed-precision GenPTQ variant. We did not
compare against quantization-aware training, ASR-specific
fine-tuning, or per-backbone hyperparameter search; these are
substantially more expensive and orthogonal to a calibration-time
intervention. Reported deployment numbers (\cref{app:efficiency})
are shared with any W$4$G$128$ baseline and do not isolate a
\TARQ-specific runtime contribution.

%% file: sections/A_appendix.tex
\section{Datasets and Evaluation Protocol}
\label{app:datasets}

\paragraph{Evaluation datasets.}
We evaluate on six standard English ASR corpora and two entity-rich
benchmarks. \textbf{LibriSpeech-clean / -other}~\cite{panayotov2015librispeech}
are read audiobook splits, with the \emph{other} split intentionally
noisier. \textbf{SPGI}~\cite{oneil2021spgispeech} is SPGISpeech,
professionally transcribed corporate earnings calls (mixed read and
spontaneous speech). \textbf{VoxPopuli}~\cite{wang-etal-2021-voxpopuli}
is European Parliament event speech (English split); transcripts are
parliamentary and rich in proper names and numerals.
\textbf{GigaSpeech}~\cite{chen2021gigaspeech} is a 10k-hour multi-domain
corpus drawn from audiobooks, podcasts, and YouTube.
\textbf{TED-LIUM}~\cite{hernandez2018ted} is TED-talk speech.
\textbf{ProfASR}~\cite{piskala2025profasrbench} targets professional
domain speech (lectures and technical talks).
\textbf{ContextASR-Speech-En}~\cite{wang2025asrbench} is a
contextual-entity-enriched ASR benchmark in which references contain a
high density of named entities.

\paragraph{Test-set subsampling.}
For each evaluation dataset, we randomly subsample $3{,}000$ utterances
from the official test split (or use the full split if it contains
fewer than $3{,}000$; only VoxPopuli-en/test falls in this case, with
$1{,}842$ utterances). The same subsample is shared across all PTQ
methods and all backbones, so cross-method WER and rare-WER comparisons
are paired on identical references. This subsample keeps decoding time
tractable on the eight backbones $\times$ six datasets $\times$ five
PTQ methods grid while remaining large enough that single-token
fluctuations do not affect rank ordering.

\paragraph{Calibration corpora.}
Calibration corpora are drawn from the training-side splits of
LibriSpeech, SPGI, and VoxPopuli. The three rare-biased pools
(\texttt{r-top}, \texttt{r-mix}, \texttt{r-cross}) are constructed by
rare-density sampling; \cref{app:cross} gives the exact rule.

\section{Model Sizes, Compute Budget, and Infrastructure}
\label{app:compute}

\paragraph{Model sizes.}
We use eight ASR backbones with the following parameter counts:
Whisper-tiny ($39$M), Whisper-base ($74$M), Whisper-small ($244$M),
Whisper-medium ($769$M), Whisper-large-v3 ($1.55$B),
Qwen3-ASR-$0.6$B ($600$M), Qwen3-ASR-$1.7$B ($1.7$B), and
Voxtral-Mini-$3$B-$2507$ ($3$B). All checkpoints are the public
release weights from the corresponding model cards.

\paragraph{Calibration cost.}
\TARQ\ is a one-pass post-training procedure. Each
(backbone, calibration corpus, PTQ method) cell quantizes the
Linear-layer weights from $128$ calibration utterances and stores
the second-moment statistics in fp32; no gradient updates and no
QAT-style fine-tuning are performed. End-to-end calibration takes
under $5$ minutes for the Whisper backbones, $\sim$$15$ minutes for
Qwen3-ASR, and $\sim$$30$ minutes for Voxtral-Mini-3B on a single
NVIDIA A100-80GB.

\paragraph{Evaluation cost.}
The main grid is $8$ backbones $\times$ $6$ datasets $\times$ $5$ PTQ
methods $\times$ $3{,}000$ utterances per test split, plus the four
entity-rich cells (\cref{sec:exp-entity-bench}) and the
cross-calibration sweep ($4$ backbones $\times$ $6$ datasets $\times$
$6$ calibration corpora). Decoding consumes the majority of compute,
and was distributed across $4{\times}$ A100-80GB GPUs over roughly
two weeks of wall time.

\paragraph{Efficiency benchmarks.}
The latency and memory benchmarks in \cref{app:efficiency} use a
single NVIDIA A100-80GB (GPU rows; Whisper / Qwen via
\texttt{whisper.cpp} / \texttt{qwen3-asr.cpp} ggml-CUDA Q4\_0,
Voxtral via Marlin INT4 in PyTorch with \texttt{gptqmodel} v5.8)
and an AMD EPYC 7V12 server CPU (CPU rows; \texttt{whisper.cpp}
and \texttt{qwen3-asr.cpp} at $\{4,8,16\}$ OpenMP threads).

\paragraph{Software.}
PyTorch $2.7$ with CUDA $12.8$, \texttt{transformers} $4.57$,
\texttt{gptqmodel} v$5.8$ (Marlin INT4 GEMM kernel for the GPU
efficiency table), \texttt{wordfreq} $3.0$+ (English Zipf scoring),
\texttt{jiwer} (default tokenization, for WER), \texttt{whisper.cpp}
commit pin in the release artifact, \texttt{qwen3-asr.cpp} after a
tensor-shape patch to the GPU-offload loader (without the patch the
upstream GPU path crashes on Qwen-1.7B; patch included in our code
release).

\paragraph{Artifact licenses.}
All public artifacts used in this work are consumed under their
released licenses; we do not redistribute them. Models:
Whisper (MIT), Voxtral-Mini-3B (Apache 2.0), Qwen3-ASR
(model-card release license, research/commercial use).
PTQ baselines and tools: \GPTQ\ / \GPTQ-style \texttt{gptqmodel}
(Apache 2.0), \AWQ\ (MIT), \OmniQuant\ (research license),
\SmoothQuant\ (MIT), \SpQR\ (Apache 2.0). Datasets:
LibriSpeech (CC~BY 4.0), VoxPopuli (CC0), GigaSpeech (Apache 2.0),
TED-LIUM (CC~BY-NC-ND 3.0), SPGISpeech (research license from
Kensho, used for non-commercial calibration / evaluation only),
ProfASR and ContextASR-Speech-En (released under research licenses
per their model cards). Inference runtimes: \texttt{whisper.cpp}
(MIT), \texttt{qwen3-asr.cpp} (MIT), \texttt{wordfreq} (MIT). Our
use is consistent with each artifact's intended research use; any
patches we contribute (e.g., the \texttt{qwen3-asr.cpp} GPU-offload
tensor-shape fix) are released under the original project's license.

\section{Background: GPTQ}
\label{app:gptq-bg}

For self-containment we recall the \GPTQ~\cite{gptq} procedure that
underlies the host solver used by \TARQ\ in \cref{sec:method-pipeline}.
\GPTQ\ quantizes a Linear-layer weight matrix $W\in\R^{m\times n}$
one column at a time, propagating each column's quantization error
to the remaining un-quantized columns under an inverse second-moment
metric. Let $H$ be the (damped) per-Linear-layer second moment from
\cref{sec:diag-defs} and let $L$ be the upper-triangular Cholesky
factor of $H^{-1}$, i.e., $H^{-1}=LL^\top$. \Cref{alg:gptq} gives
the resulting column sweep.

\begin{algorithm}[t]
\small
\DontPrintSemicolon
\SetKwInOut{KwIn}{Input}
\SetKwInOut{KwOut}{Output}
\SetKwComment{Comment}{\quad$\triangleright$\ }{}

\KwIn{Weight $W\in\R^{m\times n}$; metric $H$; quantization grid
$\mathrm{Quantize}(\cdot;b,g)$ with bit-width $b$ and group size $g$.}
\KwOut{Quantized weight $\widehat W\in\mathcal{Q}^{m\times n}$.}

Compute $H^{-1}$ and its Cholesky factor $L$ with $H^{-1}=LL^\top$\;
$\widetilde W \leftarrow W$\;
\For{$j = 1$ \KwTo $n$}{
  $\widehat w_j \leftarrow \mathrm{Quantize}(\widetilde w_j;b,g)$
  \Comment*[r]{column quantize}
  $e_j \leftarrow (\widetilde w_j - \widehat w_j)/L_{jj}$
  \Comment*[r]{normalized error}
  $\widetilde W_{:,\,j+1:n} \leftarrow \widetilde W_{:,\,j+1:n}
   - e_j L_{j,\,j+1:n}$
  \Comment*[r]{propagate}
}
\Return $\widehat W = [\widehat w_1, \ldots, \widehat w_n]$\;

\caption{\GPTQ~\cite{gptq}: column-sweep weight quantization under
metric $H$. \TARQ\ uses this routine with $H$ replaced by
$H^{\mathrm{rB}}_\ell$ (\cref{eq:rarebal}).}
\label{alg:gptq}
\end{algorithm}

\section{Reweighting Source and Cost-Ratio Ablations}
\label{app:weighting-and-c-sweep}

\paragraph{Which positions matter.}
\Cref{tab:weighting-ablation} probes \emph{which} positions the
metric-consistent solver should upweight: rare (\rareBAL),
size-matched random (\textbf{nB}, noise control), or common
(\textbf{cB}, inverse). \rareBAL\ is the unambiguous winner on both
backbones and is the only configuration that improves over \GPTQ\ on
Q-0.6B; the random and common controls drift up by $+0.4$ to $+1.8$\,pp
rare-WER. The rarity signal is therefore not just any reweighting---
only upweighting the lexical tail recovers the gains reported in
\cref{tab:lsclean-compact}.

\begin{table}[!htbp]
\centering\small
\caption{\textbf{Reweighting source ablation.} For each backbone we
upweight three position sets---rare (\textbf{rB}, \TARQ's choice),
size-matched random (\textbf{nB}, noise control), common
(\textbf{cB}, inverse)---each paired with the metric-consistent
residual correction. $\Delta$R: rare-WER change vs GPTQ. \TARQ\ row
shaded.}
\label{tab:weighting-ablation}
\setlength{\tabcolsep}{4pt}
\renewcommand{\arraystretch}{0.95}
\footnotesize
\begin{tabular}{ll rr r}
\toprule
Model & Method & Plain & Rare & $\Delta$R \\
\midrule
\multirow{4}{*}{W-b}
& GPTQ                                  & $9.48$ & $47.57$ & --- \\
& \cellcolor{gray!15}\textbf{rB + res. (\TARQ)} & \cellcolor{gray!15}$9.31$ & \cellcolor{gray!15}$\mathbf{43.96}$ & \cellcolor{gray!15}$\mathbf{-3.61}$ \\
& nB + res.                             & $9.06$ & $45.28$ & $-2.29$ \\
& cB + res.                             & $9.17$ & $44.72$ & $-2.85$ \\
\midrule
\multirow{4}{*}{Q-0.6B}
& GPTQ                                  & $6.76$ & $34.11$ & --- \\
& \cellcolor{gray!15}\textbf{rB + res. (\TARQ)} & \cellcolor{gray!15}$6.57$ & \cellcolor{gray!15}$\mathbf{32.78}$ & \cellcolor{gray!15}$\mathbf{-1.33}$ \\
& nB + res.                             & $6.58$ & $34.50$ & $+0.39$ \\
& cB + res.                             & $6.91$ & $35.86$ & $+1.75$ \\
\bottomrule
\end{tabular}
\end{table}

\paragraph{Cost-ratio default $c=1$.}
\rareBAL\ sets $\lambda_\ell$ by trace equalization, corresponding
to $c=1$ in the family $\lambda_\ell^{(c)} = c\cdot
\mathrm{tr}(H^{\common}_\ell)/\mathrm{tr}(H^{\tail}_\ell)$
(\cref{sec:method-rarity}). Sweeping $c\in\{0.25, 0.5, 1, 2, 4\}$
on three backbones (\cref{tab:c-sweep}) shows the default is not
fragile: rare-WER varies by at most $1.7$\,pp across the $16\times$
range on Whisper backbones. On Qwen3-ASR-0.6B, $c=1$ is the best
choice by a clear margin, with off-balance values clustering
$\sim$$3$\,pp higher. Trace equalization is a robust default and,
on Q-0.6B, a precise sweet spot.

\begin{table}[!htbp]
\centering\small
\caption{\textbf{Cost-ratio sweep.} Mean rare-WER (\%) across six
datasets under LS-clean calibration as a function of cost ratio
$c$. $c=1$ is the trace-equalization default used throughout the
paper. Bold: best $c$ per backbone.}
\label{tab:c-sweep}
\setlength{\tabcolsep}{5pt}
\begin{tabular}{lrr >{\columncolor{gray!15}}r rr}
\toprule
Backbone & $c=0.25$ & $c=0.5$ & $c=1$ & $c=2$ & $c=4$ \\
\midrule
W-t      & 55.3 & 55.7 & 55.93 & \textbf{54.5} & 56.2 \\
W-b      & 45.2 & 44.0 & \textbf{43.96} & 44.5 & 45.0 \\
Q-0.6B   & 36.0 & 35.8 & \textbf{32.78} & 35.5 & 35.5 \\
\bottomrule
\end{tabular}
\end{table}

\section{Zipf-Threshold Robustness}
\label{app:zipf_robustness}

The \rareBAL\ coefficient $\lambda_\ell$ depends on the Zipf threshold
$k_c$ that defines the tail partition. \Cref{tab:zipf-robust} sweeps
$k_c\in\{2,3,4\}$ at calibration and \emph{independently} re-scores
rare-WER at $k_e\in\{2,3,4\}$ at evaluation; off-diagonal cells
($k_c\neq k_e$) contain no oracle leakage between calibration and
evaluation partitions.

\begin{table*}[!htbp]
\centering
\caption{\textbf{Zipf-threshold robustness} on W-base and Q-$0.6$B. Calibration uses \rareBAL\ with Zipf$<\!k_c$ as the rare-token partition; evaluation re-scores rare-WER at Zipf$<\!k_e$ for $k_c, k_e \in \{2, 3, 4\}$. Each cell shows plain WER / rare-WER (\%) per dataset. \textbf{Bold $k_e$} marks the matched ($k_c{=}k_e$) diagonal; off-diagonal rows test cross-threshold transfer (no oracle leakage). In the \textbf{Mean} column, the lowest rare-WER across the three $k_c$ choices (for each fixed $k_e$) is bolded.}
\label{tab:zipf-robust}
\setlength{\tabcolsep}{3pt}
\renewcommand{\arraystretch}{0.95}
\footnotesize
\begin{tabular}{lcc *{7}{rr}}
\toprule
& & & \multicolumn{2}{c}{LS-c} & \multicolumn{2}{c}{LS-o} & \multicolumn{2}{c}{SPGI} & \multicolumn{2}{c}{VoxP} & \multicolumn{2}{c}{Giga} & \multicolumn{2}{c}{TedL} & \multicolumn{2}{c}{\textbf{Mean}} \\
\cmidrule(lr){4-5}\cmidrule(lr){6-7}\cmidrule(lr){8-9}\cmidrule(lr){10-11}\cmidrule(lr){12-13}\cmidrule(lr){14-15}\cmidrule(lr){16-17}
Model & $k_c$ & $k_e$ & P & R & P & R & P & R & P & R & P & R & P & R & P & R \\
\midrule
W-b & 2 & \textbf{2} & 5.50 & 65.13 & 12.78 & 77.91 & 5.83 & 50.47 & 11.27 & 57.56 & 13.01 & 40.94 & 6.44 & 40.37 & 9.14 & \textbf{55.40} \\
 &  & 3 & 5.50 & 40.85 & 12.78 & 61.10 & 5.83 & 40.30 & 11.27 & 48.75 & 13.01 & 41.32 & 6.44 & 35.24 & 9.14 & 44.59 \\
 &  & 4 & 5.50 & 20.24 & 12.78 & 38.21 & 5.83 & 18.18 & 11.27 & 23.79 & 13.01 & 25.73 & 6.44 & 15.70 & 9.14 & \textbf{23.64} \\
\cmidrule(lr){2-17}
 & 3 & 2 & 5.39 & 64.16 & 13.38 & 78.66 & 6.16 & 53.85 & 11.31 & 57.56 & 13.10 & 39.50 & 6.49 & 44.81 & 9.31 & 56.42 \\
 &  & \textbf{3} & 5.39 & 40.03 & 13.38 & 62.29 & 6.16 & 40.39 & 11.31 & 47.18 & 13.10 & 39.95 & 6.49 & 33.95 & 9.31 & \textbf{43.96} \\
 &  & 4 & 5.39 & 20.03 & 13.38 & 38.55 & 6.16 & 18.69 & 11.31 & 23.97 & 13.10 & 25.58 & 6.49 & 16.44 & 9.31 & 23.88 \\
\cmidrule(lr){2-17}
 & 4 & 2 & 5.57 & 65.27 & 12.83 & 77.59 & 5.95 & 57.41 & 11.29 & 55.70 & 12.96 & 39.65 & 6.55 & 48.52 & 9.19 & 57.36 \\
 &  & 3 & 5.57 & 41.07 & 12.83 & 61.10 & 5.95 & 43.00 & 11.29 & 48.10 & 12.96 & 40.98 & 6.55 & 39.48 & 9.19 & 45.62 \\
 &  & \textbf{4} & 5.57 & 20.41 & 12.83 & 38.53 & 5.95 & 19.28 & 11.29 & 23.54 & 12.96 & 25.76 & 6.55 & 17.06 & 9.19 & 24.10 \\
\midrule
Q-0.6B & 2 & \textbf{2} & 3.19 & 50.07 & 5.87 & 68.04 & 4.38 & 48.57 & 9.89 & 55.11 & 11.27 & 52.32 & 5.24 & 50.68 & 6.64 & 54.13 \\
 &  & 3 & 3.19 & 24.77 & 5.87 & 41.66 & 4.38 & 30.23 & 9.89 & 35.73 & 11.27 & 38.27 & 5.24 & 27.66 & 6.64 & 33.05 \\
 &  & 4 & 3.19 & 11.71 & 5.87 & 22.24 & 4.38 & 13.63 & 9.89 & 16.92 & 11.27 & 21.15 & 5.24 & 11.44 & 6.64 & \textbf{16.18} \\
\cmidrule(lr){2-17}
 & 3 & 2 & 3.16 & 52.72 & 5.75 & 69.33 & 4.04 & 47.76 & 9.95 & 54.55 & 11.21 & 49.12 & 5.29 & 50.68 & 6.57 & \textbf{54.03} \\
 &  & \textbf{3} & 3.16 & 25.96 & 5.75 & 40.58 & 4.04 & 24.80 & 9.95 & 39.23 & 11.21 & 40.21 & 5.29 & 25.89 & 6.57 & \textbf{32.78} \\
 &  & 4 & 3.16 & 12.19 & 5.75 & 22.44 & 4.04 & 13.26 & 9.95 & 18.38 & 11.21 & 21.38 & 5.29 & 11.02 & 6.57 & 16.44 \\
\cmidrule(lr){2-17}
 & 4 & 2 & 3.38 & 51.54 & 6.21 & 68.39 & 4.32 & 44.22 & 10.14 & 55.68 & 11.72 & 59.96 & 5.29 & 53.42 & 6.84 & 55.54 \\
 &  & 3 & 3.38 & 25.96 & 6.21 & 42.22 & 4.32 & 28.19 & 10.14 & 38.56 & 11.72 & 41.39 & 5.29 & 27.16 & 6.84 & 33.91 \\
 &  & \textbf{4} & 3.38 & 12.32 & 6.21 & 22.97 & 4.32 & 12.70 & 10.14 & 18.74 & 11.72 & 23.73 & 5.29 & 10.91 & 6.84 & 16.90 \\
\bottomrule
\end{tabular}
\end{table*}
Three observations support that \TARQ's mechanism is not tied to a
specific threshold. (i) Along any column (fixed $k_e$), varying $k_c$
shifts mean rare-WER by at most $\sim$2\,pp on W-b and $\sim$1.5\,pp
on Q-0.6B---small relative to the rare-WER improvements over \GPTQ\ in
\cref{tab:lsclean-compact}. (ii) Off-diagonal cells remain competitive
with their matched-diagonal counterparts; the matched diagonal
$(k_c=k_e)$ is best only at $k_e=3$, while at $k_e=2$ and $k_e=4$ the
narrower calibration partition $k_c=2$ generalizes equally well or
better, indicating that the calibration partition transfers across
evaluation partitions rather than overfitting to one. (iii) Plain WER
is essentially flat across $k_c$ (within $0.2$\,pp on W-b and
$0.3$\,pp on Q-0.6B), so the threshold choice does not trade aggregate
accuracy for rare-token accuracy. Notably, $k_c=4$ is never optimal
under any $(k_e,\text{backbone})$ combination, suggesting that
\rareBAL\ requires the partition to genuinely isolate low-frequency
tokens: broadening the tail definition to include mid-frequency
vocabulary (Zipf $\in[3,4)$) dilutes the rebalancing signal. Together,
these observations indicate that \rareBAL\ captures a general
common/tail imbalance rather than a Zipf-3-specific signal.

\section{Integration with Other PTQ Methods}
\label{app:integration}

\rareBAL\ changes only the calibration metric, so it can be grafted
onto other PTQ pipelines whose host solver consumes a Hessian-like
matrix. We test two integrations: \SmoothQuant~\cite{xiao2023smoothquant},
which applies a per-channel activation--weight rescaling before
quantization, and \SpQR~\cite{dettmers2023spqr}, which keeps a small
fraction of weights at higher precision. \Cref{tab:sq-graft-per-dataset}
reports the \SmoothQuant\ graft and \cref{tab:spqr-graft-per-dataset}
the \SpQR\ graft.

\subsection{\SmoothQuant\ graft}
\label{app:integration-sq}

We replace \GPTQ\ with \TARQ\ as the per-Linear-layer solver inside
\SmoothQuant, leaving the activation-rescaling stage unchanged.

\begin{table*}[!htbp]
\centering
\caption{\textbf{\TARQ\ as a graft on SmoothQuant}: per-dataset plain~WER / rare-WER (\%) under W4G128 weight-only (W4A16) and original-SmoothQuant W4A8 (per-token dynamic INT8 activation). Bold = best, underline = second-best per (Backbone, dataset, metric).}
\label{tab:sq-graft-per-dataset}
\setlength{\tabcolsep}{2pt}
\renewcommand{\arraystretch}{0.9}
\small
\begin{tabular}{ll *{7}{rr}}
\toprule
& & \multicolumn{2}{c}{LSc} & \multicolumn{2}{c}{LSo} & \multicolumn{2}{c}{SPGI} & \multicolumn{2}{c}{VoxP} & \multicolumn{2}{c}{Giga} & \multicolumn{2}{c}{TedL} & \multicolumn{2}{c}{\textbf{Mean}} \\
\cmidrule(lr){3-4}\cmidrule(lr){5-6}\cmidrule(lr){7-8}\cmidrule(lr){9-10}\cmidrule(lr){11-12}\cmidrule(lr){13-14}\cmidrule(lr){15-16}
Backbone & Method & P & R & P & R & P & R & P & R & P & R & P & R & P & R \\
\midrule
\multirow{2}{*}{W-t (W4A16)}
& SQ+GPTQ & \underline{9.95} & \underline{13.64} & \underline{21.43} & \underline{28.00} & \underline{12.29} & \underline{12.99} & \underline{15.62} & \underline{15.64} & \underline{17.22} & \underline{15.84} & \underline{10.12} & \underline{9.22} & \underline{14.44} & \underline{15.89} \\
& \cellcolor{gray!15}SQ+\TARQ & \cellcolor{gray!15}\textbf{9.28} & \cellcolor{gray!15}\textbf{12.89} & \cellcolor{gray!15}\textbf{19.77} & \cellcolor{gray!15}\textbf{26.14} & \cellcolor{gray!15}\textbf{11.23} & \cellcolor{gray!15}\textbf{11.86} & \cellcolor{gray!15}\textbf{15.18} & \cellcolor{gray!15}\textbf{15.18} & \cellcolor{gray!15}\textbf{16.54} & \cellcolor{gray!15}\textbf{15.18} & \cellcolor{gray!15}\textbf{9.85} & \cellcolor{gray!15}\textbf{9.00} & \cellcolor{gray!15}\textbf{13.64} & \cellcolor{gray!15}\textbf{15.04} \\
\midrule
\multirow{2}{*}{W-t (W4A8)}
& SQ+GPTQ & \underline{10.11} & \underline{13.91} & \underline{21.31} & \underline{27.91} & \underline{12.43} & \underline{13.09} & \underline{15.62} & \underline{15.58} & \underline{17.35} & \underline{16.03} & \underline{10.47} & \underline{9.74} & \underline{14.55} & \underline{16.04} \\
& \cellcolor{gray!15}SQ+\TARQ & \cellcolor{gray!15}\textbf{9.41} & \cellcolor{gray!15}\textbf{13.14} & \cellcolor{gray!15}\textbf{20.11} & \cellcolor{gray!15}\textbf{26.49} & \cellcolor{gray!15}\textbf{11.33} & \cellcolor{gray!15}\textbf{11.90} & \cellcolor{gray!15}\textbf{15.29} & \cellcolor{gray!15}\textbf{15.27} & \cellcolor{gray!15}\textbf{16.75} & \cellcolor{gray!15}\textbf{15.40} & \cellcolor{gray!15}\textbf{9.93} & \cellcolor{gray!15}\textbf{9.12} & \cellcolor{gray!15}\textbf{13.80} & \cellcolor{gray!15}\textbf{15.22} \\
\midrule
\multirow{2}{*}{W-b (W4A16)}
& SQ+GPTQ & \underline{6.07} & \underline{8.59} & \underline{13.67} & \underline{18.79} & \underline{6.42} & \underline{6.74} & \underline{11.74} & \underline{11.18} & \underline{13.42} & \underline{11.65} & \underline{6.70} & \underline{5.81} & \underline{9.67} & \underline{10.46} \\
& \cellcolor{gray!15}SQ+\TARQ & \cellcolor{gray!15}\textbf{5.74} & \cellcolor{gray!15}\textbf{8.24} & \cellcolor{gray!15}\textbf{13.10} & \cellcolor{gray!15}\textbf{18.11} & \cellcolor{gray!15}\textbf{6.23} & \cellcolor{gray!15}\textbf{6.52} & \cellcolor{gray!15}\textbf{11.45} & \cellcolor{gray!15}\textbf{10.86} & \cellcolor{gray!15}\textbf{13.20} & \cellcolor{gray!15}\textbf{11.42} & \cellcolor{gray!15}\textbf{6.62} & \cellcolor{gray!15}\textbf{5.59} & \cellcolor{gray!15}\textbf{9.39} & \cellcolor{gray!15}\textbf{10.12} \\
\midrule
\multirow{2}{*}{W-b (W4A8)}
& SQ+GPTQ & \underline{6.08} & \underline{8.58} & \underline{13.74} & \underline{18.91} & \underline{6.42} & \underline{6.72} & \underline{11.71} & \underline{11.16} & \underline{13.63} & \underline{11.80} & \underline{6.84} & \underline{5.89} & \underline{9.74} & \underline{10.51} \\
& \cellcolor{gray!15}SQ+\TARQ & \cellcolor{gray!15}\textbf{5.77} & \cellcolor{gray!15}\textbf{8.27} & \cellcolor{gray!15}\textbf{13.25} & \cellcolor{gray!15}\textbf{18.20} & \cellcolor{gray!15}\textbf{6.29} & \cellcolor{gray!15}\textbf{6.56} & \cellcolor{gray!15}\textbf{11.48} & \cellcolor{gray!15}\textbf{10.92} & \cellcolor{gray!15}\textbf{13.26} & \cellcolor{gray!15}\textbf{11.52} & \cellcolor{gray!15}\textbf{6.67} & \cellcolor{gray!15}\textbf{5.71} & \cellcolor{gray!15}\textbf{9.45} & \cellcolor{gray!15}\textbf{10.20} \\
\midrule
\multirow{2}{*}{Q-0.6B}
& SQ+GPTQ & \underline{2.98} & \underline{4.70} & \underline{5.57} & \underline{8.48} & \textbf{4.39} & \underline{5.29} & \textbf{9.84} & \underline{9.25} & \textbf{11.03} & \textbf{9.76} & \textbf{4.50} & \textbf{4.21} & \textbf{6.39} & \textbf{6.95} \\
& \cellcolor{gray!15}SQ+\TARQ & \cellcolor{gray!15}\textbf{2.92} & \cellcolor{gray!15}\textbf{4.54} & \cellcolor{gray!15}\textbf{5.52} & \cellcolor{gray!15}\textbf{8.32} & \cellcolor{gray!15}\underline{4.55} & \cellcolor{gray!15}\textbf{5.28} & \cellcolor{gray!15}\underline{10.04} & \cellcolor{gray!15}\textbf{9.20} & \cellcolor{gray!15}\underline{11.48} & \cellcolor{gray!15}\underline{10.44} & \cellcolor{gray!15}\underline{5.11} & \cellcolor{gray!15}\underline{5.25} & \cellcolor{gray!15}\underline{6.60} & \cellcolor{gray!15}\underline{7.17} \\
\bottomrule
\end{tabular}
\end{table*}
\paragraph{Whisper-tiny/-base.}
Grafting \TARQ\ onto SmoothQuant improves both plain and rare~WER on
every dataset, under both W4A16 and W4A8. Mean plain~WER drops by
$0.75$--$0.80$~pp on W-t and $\sim$$0.28$~pp on W-b; rare~WER drops by
a similar margin ($0.82$--$0.85$~pp on W-t, $\sim$$0.32$~pp on W-b).
The W4A16$\leftrightarrow$W4A8 gap is small for both methods
($\leq$$0.16$~pp in mean plain~WER), so the rebalancing gain is
largely orthogonal to activation precision.

\paragraph{Qwen3-ASR-0.6B.}
The picture is mixed. \TARQ\ wins on the two LibriSpeech splits (lower
plain and rare~WER) but regresses on SPGI, VoxPopuli, GigaSpeech, and
TED-LIUM, leaving mean plain~WER $0.21$~pp and mean rare~WER
$0.22$~pp behind SQ+GPTQ. We attribute this to LS-clean calibration:
the Zipf statistics estimated from clean read speech transfer well to
LS-other but less well to the other four domains, and on a stronger
backbone the residual headroom is small enough for that mismatch to
dominate. The Whisper results, where the same LS-clean calibration
helps uniformly, suggest the issue is calibration--evaluation domain
shift rather than the reweighting itself; we revisit this in the
Limitations section (\cref{sec:limitations}).

\subsection{\SpQR\ graft with orthogonality gate}
\label{app:integration-spqr}

\SpQR\ keeps a small fraction of weights at FP16 to absorb outliers.
A naive \SpQR{}+\TARQ\ graft produces a coverage conflict on
Q-$0.6$B: the FP16 outliers and \rareBAL's rarity weights target the
same high-leverage positions, so the rare-token budget gets
double-counted. The \texttt{--rarity-gate-outliers} flag
(\cref{alg:spqr-parq-gate}) zeroes rarity weights on tokens already
covered by \SpQR\ outliers; \cref{tab:spqr-graft-per-dataset} reports
the per-dataset effect.

\begin{table*}[!htbp]
\centering
\caption{\textbf{\TARQ\ + \SpQR\ with orthogonality gate}: per-dataset plain~WER / rare-WER (\%) at W4G128 + $1\%$ outlier-FP16. Naive \SpQR\,+\,\TARQ\ shows coverage conflict (Q-$0.6$B regresses $+0.37$~pp rare); the gate restores positive $\Delta$. Bold: rank-1 per (backbone, dataset, metric); underline: rank-2.}
\label{tab:spqr-graft-per-dataset}
\setlength{\tabcolsep}{2pt}
\renewcommand{\arraystretch}{0.9}
\small
\begin{tabular}{ll *{7}{rr}}
\toprule
& & \multicolumn{2}{c}{LSc} & \multicolumn{2}{c}{LSo} & \multicolumn{2}{c}{SPGI} & \multicolumn{2}{c}{VoxP} & \multicolumn{2}{c}{Giga} & \multicolumn{2}{c}{TedL} & \multicolumn{2}{c}{\textbf{Mean}} \\
\cmidrule(lr){3-4}\cmidrule(lr){5-6}\cmidrule(lr){7-8}\cmidrule(lr){9-10}\cmidrule(lr){11-12}\cmidrule(lr){13-14}\cmidrule(lr){15-16}
Backbone & Method & P & R & P & R & P & R & P & R & P & R & P & R & P & R \\
\midrule
\multirow{3}{*}{W-t}
& SpQR & \underline{9.09} & \textbf{12.56} & 19.48 & \underline{25.76} & 10.32 & \underline{10.61} & 14.97 & \underline{14.71} & \underline{16.35} & \underline{15.19} & \textbf{9.63} & \textbf{8.67} & 13.31 & \textbf{14.58} \\
& SpQR+\TARQ & 9.11 & 12.73 & \textbf{19.06} & \textbf{25.45} & \underline{10.22} & 10.74 & \underline{14.88} & 14.75 & 16.38 & 15.28 & 10.04 & 9.48 & \underline{13.28} & 14.74 \\
\rowcolor{gray!15}
& SpQR+\TARQ+gate & \textbf{8.95} & \underline{12.59} & \underline{19.36} & 25.95 & \textbf{10.05} & \textbf{10.38} & \textbf{14.86} & \textbf{14.66} & \textbf{16.21} & \textbf{15.01} & \underline{9.84} & \underline{9.23} & \textbf{13.21} & \underline{14.64} \\
\midrule
\multirow{3}{*}{W-b}
& SpQR & 5.79 & 8.33 & \underline{13.12} & 18.13 & 6.05 & 6.27 & 11.55 & 10.92 & 13.11 & 11.26 & \underline{6.54} & 5.41 & 9.36 & 10.05 \\
& SpQR+\TARQ & \underline{5.66} & \textbf{8.16} & \textbf{12.83} & \underline{17.68} & \underline{5.89} & \underline{6.01} & \underline{11.35} & \underline{10.77} & \textbf{12.95} & \underline{11.12} & \textbf{6.52} & \underline{5.33} & \underline{9.20} & \underline{9.85} \\
\rowcolor{gray!15}
& SpQR+\TARQ+gate & \textbf{5.63} & \underline{8.22} & \textbf{12.83} & \textbf{17.58} & \textbf{5.85} & \textbf{5.97} & \textbf{11.14} & \textbf{10.52} & \underline{13.01} & \textbf{11.08} & 6.62 & \textbf{5.26} & \textbf{9.18} & \textbf{9.77} \\
\midrule
\multirow{3}{*}{Q-0.6B}
& SpQR & \textbf{2.87} & \textbf{4.52} & \textbf{5.42} & \textbf{8.17} & \textbf{4.46} & \textbf{5.35} & \underline{10.11} & 9.51 & \underline{11.37} & \underline{10.20} & \textbf{4.48} & \textbf{4.31} & \textbf{6.45} & \textbf{7.01} \\
& SpQR+\TARQ & 3.11 & 4.92 & 5.87 & 8.88 & \underline{4.49} & \textbf{5.35} & 10.18 & \underline{9.45} & 11.79 & 10.74 & 5.08 & 4.95 & 6.75 & 7.38 \\
\rowcolor{gray!15}
& SpQR+\TARQ+gate & \underline{3.01} & \underline{4.76} & \underline{5.65} & \underline{8.56} & 4.58 & \underline{5.36} & \textbf{9.90} & \textbf{9.29} & \textbf{11.36} & \textbf{10.12} & \underline{4.70} & \underline{4.57} & \underline{6.53} & \underline{7.11} \\
\bottomrule
\end{tabular}
\end{table*}

\paragraph{Whisper backbones.}
On W-t and W-b, SpQR+\TARQ\ already improves mean WER over SpQR
(W-b: $9.36$/$10.05 \to 9.20$/$9.85$), and the orthogonality gate gives
a further small gain ($\to 9.18$/$9.77$ on W-b; $\to 13.21$/$14.64$ on
W-t). The W-t case is uneven at the per-dataset level---naive graft
slightly regresses rare~WER on LSc and Giga, both of which the gate
partially recovers---but mean rare~WER stays within $0.06$~pp of SpQR
alone in the worst case and improves elsewhere.

\paragraph{Qwen3-ASR-0.6B.}
Here SpQR is already a strong baseline and the interaction with
\TARQ\ becomes visible. Naive SpQR+\TARQ\ regresses on every dataset
(mean $+0.30$/$+0.37$~pp), consistent with coverage conflict: SpQR's
FP16 outliers and \TARQ's rarity weights both target the same
high-leverage positions, so rare-token reweighting double-counts the
positions SpQR has already protected and dilutes budget on the rest.
The gate, which zeroes rarity weights on tokens already covered by
SpQR outliers, closes most of this gap (mean $6.53$/$7.11$, within
$0.08$/$0.10$~pp of SpQR) but does not recover a win on this backbone.
We read this as evidence that \TARQ\ is most useful when it has unique
coverage to add; on backbones where an outlier-protection scheme
already absorbs the rare-token budget, the gate keeps the graft safe
but the marginal headroom is small.
\begin{algorithm}[t]
\caption{\SpQR\ + \TARQ\ with orthogonality gate.}
\label{alg:spqr-parq-gate}
\KwIn{Layer $W$, calibration activations $\{x_t\}_{t=1}^{T}$, rarity weights
$\{w_t\}$, $\rho{=}0.01$, $\tau{=}3$.}
\KwOut{Quantized $\hat W$ with \TARQ\ + \SpQR\ outlier protection.}
\textbf{Pass~1.}
$\Hwz \leftarrow \sum_t x_t x_t^\top + 0.01\cdot\mathrm{tr}(\Hwz)/d \cdot I$;
$s_j \leftarrow \|W_{:,j}\|_2^2 / [\Hwz^{-1}]_{jj}$;
$M \leftarrow \{j: s_j \ge \mathrm{topk}(s, \lceil\rho d\rceil)\}$\;
\textbf{Pass~2.}
$g_t \leftarrow \mathbf{1}[\max_{j\in M}|x_{t,j}| > \tau\,\E_{t,j}|x_{t,j}|]$;
$\tilde w_t \leftarrow (1{-}g_t) w_t + g_t$;
$H \leftarrow \sum_t \tilde w_t x_t x_t^\top$\;
\textbf{Quantize.}
$\hat W \leftarrow \mathrm{SpQR}(W,H)$ \cite{dettmers2023spqr}.
\end{algorithm}
\input{sections/B_efficiency}

\section{Full Numerical Results}
\label{app:full-tables}

This section gives the per-dataset numbers behind the main-paper
aggregate tables. \Cref{tab:cross-calib-summary}
(\cref{app:cross}) reports the cross-corpus rare-WER sweep: \TARQ\ is
rank-1 on $5$ of $6$ calibration corpora and has the smallest
cross-corpus swing among compared fixed-hyperparameter baselines. \Cref{tab:ablation-full}
(\cref{app:ablation_full}) decomposes \TARQ\ into \rareBAL\ and the
residual correction across all six evaluation datasets, confirming
that \rareBAL\ alone closes most of the rare-WER gap on both ablated
backbones. \Cref{tab:lsclean-full} (\cref{app:full-table1}) gives the
per-dataset plain WER and rare-WER for all eight backbones under
LS-clean calibration; \TARQ\ achieves rank-1 mean plain WER on every
backbone and rank-1 mean rare-WER on six of eight, with the two
remaining cells (W-m, V-3B) won by mixed-precision GenPTQ by a thin
margin.

\subsection{Cross-dataset calibration robustness}
\label{app:cross}

\Cref{tab:cross-calib-summary} gives the per-method,
per-calibration-corpus numbers visualized in
\cref{fig:main-and-cross}(c), including the cross-corpus swing.

\paragraph{Rare-biased calibration pools.}
The three rare-biased calibration corpora (\textbf{r-top}, \textbf{r-mix},
\textbf{r-cross} in \cref{fig:main-and-cross}(c)) are constructed from
the training-side ASR corpora by scoring every utterance with its
\emph{rare-density}
\begin{equation}
\rho(\text{utt}) \;=\;
\frac{\bigl|\{w \in \text{utt}\,:\, \mathrm{Zipf}(w) < 3.0\}\bigr|}{|\text{utt}|},
\end{equation}
using the \texttt{wordfreq}~\cite{word_freq} English Zipf scale---the
same $\mathrm{Zipf}{<}3$ partition that defines the tail class at
calibration time. Given a source corpus $D$ and target calibration
size $N{=}128$:
\begin{itemize}
\item \textbf{r-top.} Take the utterances with the $N$ largest $\rho$
values from $D$ alone: $\arg\text{top-}N\,\rho(D)$. The most
aggressively rare-skewed pool: every utterance has above-average
rare-token density within its source.
\item \textbf{r-mix.} Half rare-biased, half natural: $N/2$ utterances
by $\arg\text{top-}\rho$ from $D$, plus $N/2$ utterances sampled
uniformly at random from $D \setminus S_{\text{top}}$ (no overlap with
the top half). Tests whether \rareBAL's gains require a fully-biased
pool or remain in a mixture.
\item \textbf{r-cross.} Cross-domain rare-dense pool. Form a candidate
set $\mathcal{P}$ by taking the top $\lceil 4N/3\rceil \approx 170$
rare-dense utterances from each of LibriSpeech-train.100,
SPGI-S/train, and VoxPopuli-en/train, then return the global
$\arg\text{top-}N\,\rho$ over $\mathcal{P}$. Removes single-source
bias: the resulting calibration batch draws rare-dense utterances
from three different acoustic distributions.
\end{itemize}
The three pools probe complementary axes: r-top isolates pure
rare-density within a single source, r-mix tests robustness to a
half-biased / half-natural mixture, and r-cross checks that the
rare effect is not an artifact of any one source's domain.

\input{appendix_tables/calbration_different_dataset}

\subsection{Component ablation: per-dataset breakdown}
\label{app:ablation_full}

\Cref{tab:ablation-full} expands \cref{tab:ablation} of
\cref{sec:component-analysis} with per-dataset cells.

\begin{table*}[!htbp]
\centering
\caption{\textbf{Component ablation (per-dataset breakdown)} of \TARQ\ on W-b and Q-0.6B under W4G128 weight-only quantization with LS-clean calibration. Each cell reports plain WER / rare-WER (\%). \textit{+rareBAL} applies the rebalanced metric $H^{\mathrm{rB}}_\ell$ alone; \textit{+residual} applies the residual correction alone (over the standard $H_\ell$); \TARQ\ combines both. Bold: rank-1 per (cell, metric); underline: rank-2.}
\label{tab:ablation-full}
\setlength{\tabcolsep}{3pt}
\renewcommand{\arraystretch}{0.95}
\small
\begin{tabular}{ll *{7}{rr}}
\toprule
& & \multicolumn{2}{c}{LS-c} & \multicolumn{2}{c}{LS-o} & \multicolumn{2}{c}{SPGI} & \multicolumn{2}{c}{VoxP} & \multicolumn{2}{c}{Giga} & \multicolumn{2}{c}{TedL} & \multicolumn{2}{c}{\textbf{Mean}} \\
\cmidrule(lr){3-4}\cmidrule(lr){5-6}\cmidrule(lr){7-8}\cmidrule(lr){9-10}\cmidrule(lr){11-12}\cmidrule(lr){13-14}\cmidrule(lr){15-16}
Model & Method & P & R & P & R & P & R & P & R & P & R & P & R & P & R \\
\midrule
\multirow{4}{*}{W-b}
& GPTQ & 5.60 & 40.80 & 13.19 & 61.70 & 6.02 & 41.65 & 12.48 & 61.99 & 13.71 & 42.04 & \textbf{5.89} & 37.27 & 9.48 & 47.58 \\
& +rareBAL & \underline{5.51} & \underline{40.71} & \textbf{12.84} & \textbf{61.01} & \textbf{5.77} & \underline{41.36} & \underline{11.62} & \underline{48.49} & \underline{13.44} & \underline{40.79} & \underline{6.35} & \underline{35.42} & \textbf{9.26} & \underline{44.63} \\
& +residual & 5.54 & 40.77 & \underline{12.96} & \underline{61.61} & \underline{5.87} & 41.65 & 11.73 & 48.89 & 13.57 & 41.93 & 6.61 & 35.56 & 9.38 & 45.07 \\
\rowcolor{gray!15}
& \textbf{\TARQ\ } & \textbf{5.39} & \textbf{40.03} & 13.38 & 62.29 & 6.16 & \textbf{40.39} & \textbf{11.31} & \textbf{47.18} & \textbf{13.10} & \textbf{39.95} & 6.49 & \textbf{33.95} & \underline{9.31} & \textbf{43.96} \\
\midrule
\multirow{4}{*}{Q-0.6B}
& GPTQ & 3.29 & 26.55 & 5.92 & 41.14 & 4.30 & 28.87 & 10.00 & 39.73 & 11.47 & 40.44 & 5.60 & 27.92 & 6.76 & 34.11 \\
& +rareBAL & \underline{3.24} & \underline{26.23} & \underline{5.86} & \underline{40.79} & \underline{4.09} & \underline{27.74} & \underline{9.95} & \underline{38.23} & \underline{11.43} & \textbf{38.70} & \textbf{4.79} & \textbf{25.13} & \textbf{6.56} & \underline{32.80} \\
& +residual & 3.25 & \underline{26.23} & 5.88 & 40.81 & 4.20 & 28.76 & \textbf{9.90} & \textbf{37.56} & 11.44 & \underline{39.44} & \underline{5.05} & 26.92 & 6.62 & 33.29 \\
\rowcolor{gray!15}
& \textbf{\TARQ\ } & \textbf{3.16} & \textbf{25.96} & \textbf{5.75} & \textbf{40.58} & \textbf{4.04} & \textbf{24.80} & \underline{9.95} & 39.23 & \textbf{11.21} & 40.21 & 5.29 & \underline{25.89} & \underline{6.57} & \textbf{32.78} \\
\bottomrule
\end{tabular}
\end{table*}

\subsection{Per-dataset results under LS-clean calibration}
\label{app:full-table1}

\Cref{tab:lsclean-full} expands \cref{tab:lsclean-compact} of
\cref{sec:exp-main} with per-dataset cells. Each cell reports plain
WER / rare-WER (\%) under W4G128 weight-only quantization; GenPTQ
uses mixed precision with a 4-bit average target.

\input{appendix_tables/main_table}
\section{Qualitative Examples: Rare-Word Recovery}
\label{app:qualitative}

We sample concrete utterances where \TARQ\ recovers a rare reference
word that \emph{every} other W4G128 baseline (\GPTQ, \AWQ, \OmniQuant, GenPTQ)
drops or substitutes, and that FP16 also recovers. Rare here means
\texttt{wordfreq} Zipf $<3.0$ with at least four characters. Truncated
$\sim$$10$-word windows around the rare word are shown; the rare word
is rendered as \rwWord{like this} when preserved and in
\emph{italics} when substituted. \rwOK\,/\,\rwNO\ marks whether the
target word survived.

\subsection{Whisper-tiny}

\begin{rwexamplebox}
\textbf{\texttt{mademoiselle}} \quad\textit{LS-clean}\\[2pt]
\rwRef \ldots la valliere said \rwWord{mademoiselle} de tonnay charente \ldots\\
\rwOK~\rwTARQ: \ldots lavallier said \rwWord{mademoiselle} dittinichan \ldots\\
\rwNO~\GPTQ: \ldots lavaliay said \emph{madhmozell} dittinichan \ldots\\
\rwNO~\AWQ: \ldots lavallier said \emph{madmuzel} d ane shant \ldots\\
\rwNO~\OmniQuant: \ldots lavalia said \emph{maud mousel} d aixard \ldots\\
\rwNO~GenPTQ: \ldots lavalia said \emph{mademazelle} that they should not
\end{rwexamplebox}

\begin{rwexamplebox}
\textbf{\texttt{harmonized}} \quad\textit{LS-clean}\\[2pt]
\rwRef \ldots the white is inclosed properly and \rwWord{harmonized} with the other hues \ldots\\
\rwOK~\rwTARQ: \ldots properly and \rwWord{harmonized} with the other hues \ldots\\
\rwNO~\GPTQ: \ldots not closed properly and \emph{harmonize} with the other hues \ldots\\
\rwNO~\AWQ: \ldots not closed properly and \emph{harmonize} with the other hues \ldots\\
\rwNO~\OmniQuant: \ldots not closed properly and \emph{harmonize} with the other hues \ldots\\
\rwNO~GenPTQ: \ldots enclosed properly and \emph{harmonize} with the other hues \ldots
\end{rwexamplebox}

\begin{rwexamplebox}
\textbf{\texttt{pewter}} \quad\textit{LS-clean}\\[2pt]
\rwRef \ldots she could see herself sometimes in the great round \rwWord{pewter} dishes \ldots\\
\rwOK~\rwTARQ: \ldots in the great round \rwWord{pewter} dishes \ldots\\
\rwNO~\GPTQ: \ldots in the great round \emph{buter} dishes \ldots\\
\rwNO~\AWQ: \ldots in the great round \emph{putter} dishes \ldots\\
\rwNO~\OmniQuant: \emph{[output collapses to ``headies are all over the place'']}\\
\rwNO~GenPTQ: \emph{[output degenerates to a ``ground ground ground'' loop]}
\end{rwexamplebox}

\subsection{Whisper-base}

\begin{rwexamplebox}
\textbf{\texttt{valjean}} \quad\textit{LS-other}\\[2pt]
\rwRef \ldots this was one of jean \rwWord{valjean} is gloomy talents\\
\rwOK~\rwTARQ: \ldots this was one of jean \rwWord{valjean} is gloomy talents\\
\rwNO~\GPTQ: \ldots one of jean \emph{valjeant} is gloomy talents\\
\rwNO~\AWQ: \ldots one of jean \emph{vagion} is gloomy talents\\
\rwNO~\OmniQuant: \emph{this was one of jean is gloomy talents}\\
\rwNO~GenPTQ: \ldots one of jean \emph{vageau} is gloomy talents
\end{rwexamplebox}

\begin{rwexamplebox}
\textbf{\texttt{nobilis}} \quad\textit{SPGI}\\[2pt]
\rwRef responsibility for all of \rwWord{nobilis} operating units \ldots executing \rwWord{nobilis} growth strategies\\
\rwOK~\rwTARQ: \ldots all of \emph{nobillis} \ldots executing \rwWord{nobilis} growth strategies\\
\rwNO~\GPTQ: \ldots all of \emph{nobillis} \ldots executing \emph{nobillis} growth strategies\\
\rwNO~\AWQ: \ldots all of \emph{nobiluses} \ldots executing \emph{nobiluses} growth strategies\\
\rwNO~\OmniQuant: \emph{he will have responsibility for all of the} \ldots [truncated]\\
\rwNO~GenPTQ: \ldots all of the \emph{billisers} \ldots executing the \emph{billiases} \ldots
\end{rwexamplebox}

\begin{rwexamplebox}
\textbf{\texttt{marmalades}} \quad\textit{LS-other}\\[2pt]
\rwRef \rwWord{marmalades} jams and fruit pastes are of the same nature \ldots\\
\rwOK~\rwTARQ: \rwWord{marmalades} jams and fruit paces of the same nature \ldots\\
\rwNO~\GPTQ: \emph{marmalade} jams and fruit \emph{paste} are the same nature \ldots\\
\rwNO~\AWQ: \emph{marmalade} jams and fruit \emph{pace} are the same nature \ldots\\
\rwNO~\OmniQuant: \emph{marmalade} and fruit \emph{paster}\\
\rwNO~GenPTQ: \emph{marmalade is jams and fruit paste} \ldots
\end{rwexamplebox}

\subsection{Qwen3-ASR-0.6B}

\begin{rwexamplebox}
\textbf{\texttt{merganser}} \quad\textit{LS-clean}\\[2pt]
\rwRef the \rwWord{merganser} had a crested head of iridescent green black \ldots\\
\rwOK~\rwTARQ: the \rwWord{merganser} had a crested head of iridescent greenblack \ldots\\
\rwNO~\GPTQ: the \emph{morgan sir} had a crested head \ldots\\
\rwNO~\OmniQuant: the \emph{magancer} had a crested head \ldots\\
\rwNO~GenPTQ: the \emph{morganus} had a crested head \ldots
\end{rwexamplebox}

\begin{rwexamplebox}
\textbf{\texttt{dandan}} \quad\textit{LS-other}\\[2pt]
\rwRef \ldots the sultan commanded his wazir \rwWord{dandan} call a ten days halt \ldots\\
\rwOK~\rwTARQ: \ldots his wazir \rwWord{dandan} call a ten days halt \ldots\\
\rwNO~\GPTQ: \ldots his \emph{vizier done} call at ten days halt \ldots\\
\rwNO~\OmniQuant: \ldots his \emph{wazirdandan} call a ten days halt \ldots\\
\rwNO~GenPTQ: \ldots his \emph{wazirdan khan} call at ten days halt \ldots
\end{rwexamplebox}

\begin{rwexamplebox}
\textbf{\texttt{globalisation}} \quad\textit{VoxPopuli}\\[2pt]
\rwRef \ldots the losers of \rwWord{globalisation} \ldots a crisis of \rwWord{globalisation}\\
\rwOK~\rwTARQ: \ldots losers of \rwWord{globalisation} \ldots crisis of \rwWord{globalisation}\\
\rwNO~\GPTQ: \ldots losers of \emph{globalization} \ldots crisis of \emph{globalization} \quad\textit{(spelling drift to US norm)}\\
\rwNO~\OmniQuant: \ldots losers of \emph{globalization} \ldots crisis of \emph{globalization}\\
\rwNO~GenPTQ: \ldots losers of \emph{globalization} \ldots crisis of \emph{globalization}
\end{rwexamplebox}

\begin{rwexamplebox}
\textbf{\texttt{euphranor}} \quad\textit{LS-other}\\[2pt]
\rwRef \ldots seventieth birthday of the old sculptor \rwWord{euphranor}\\
\rwOK~\rwTARQ: \ldots seventieth birthday of the old sculptor \rwWord{euphranor}\\
\rwNO~\GPTQ: \ldots seventieth birthday of the old sculptor \emph{euphraner}\\
\rwNO~\OmniQuant: \ldots seventieth birthday of the old sculptor \emph{euphraner}\\
\rwNO~GenPTQ: \ldots seventieth birthday of the old sculptor \emph{euphraner}
\end{rwexamplebox}

\begin{rwexamplebox}
\textbf{\texttt{carthusians}} \quad\textit{LS-other}\\[2pt]
\rwRef \ldots eleventh general of the \rwWord{carthusians} gave to his order this device \ldots\\
\rwOK~\rwTARQ: \ldots eleventh general of the \rwWord{carthusians} gave to his order this device \ldots\\
\rwNO~\GPTQ: \ldots eleventh general of the \emph{carthagens} gave to his order \ldots\\
\rwNO~\OmniQuant: \ldots eleventh general of the \emph{carthians} gave to his order \ldots\\
\rwNO~GenPTQ: \ldots eleventh general of the \emph{carthions} gave to his order \ldots
\end{rwexamplebox}

\section{Reproducibility and Responsible-Research Statement}
\label{app:responsible}

This section completes the ARR Responsible-NLP checklist for items not
already covered by \cref{app:datasets,app:compute} (model sizes,
hardware, compute budget, software versions, and per-dataset licenses)
and the Limitations section (\cref{sec:limitations}).

\paragraph{Baseline hyperparameters.}
All baseline PTQ methods are run from their public reference
implementations under their recommended W4G128 settings;
\cref{tab:baseline-hparams} records the non-default knobs.

\begin{table*}[!htbp]
\centering\small
\caption{Baseline PTQ hyperparameters used for all (backbone, dataset)
cells. Defaults from each method's released code are used everywhere
else.}
\label{tab:baseline-hparams}
\setlength{\tabcolsep}{5pt}
\begin{tabular}{ll}
\toprule
Method & Setting \\
\midrule
\GPTQ      & \texttt{percdamp}~$0.01$, \texttt{act\_order} off \\
\AWQ       & grid search $\alpha\in[0,1]$ in $20$ steps, block-wise reconstruction \\
\OmniQuant & LWC + LET variant, $200$ iterations, lr $5\mathrm{e}{-3}$, batch $4$ \\
GenPTQ     & $150$ iterations, lr $0.1$, target $4.0$ bits, $\lambda_p=-1$ (auto), $\lambda_{sr}=1$ \\
\bottomrule
\end{tabular}
\end{table*}

\paragraph{TARQ hyperparameters.}
\TARQ\ has only two interpretable knobs and we do not tune either
per backbone: the rare-token Zipf threshold $k=3.0$
(\cref{app:zipf_robustness} sweeps $k\in\{2,3,4\}$ and shows the
choice is not load-bearing) and the cost ratio $c=1$ in
$\lambda_\ell=c\cdot\mathrm{tr}(H^{\common}_\ell)/\mathrm{tr}(H^{\tail}_\ell)$
(\cref{app:weighting-and-c-sweep} sweeps $c\in\{0.25,0.5,1,2,4\}$).

\paragraph{Artifact release.}
We will release the \TARQ\ source code (calibration pipeline after code cleaning, \rareBAL\
Hessian rebalance, closed-form residual correction, evaluation
scripts) under the Apache-$2.0$ license, and \TARQ-quantized
W4G128 checkpoints for each backbone in \cref{app:compute} under
the original backbone's license. The release is documented with a
README covering supported backbones, the \texttt{wordfreq}
Zipf~$<3$ rare-token convention used at calibration time, the
required Python/CUDA environment, and reproduction instructions for
every table in this paper. Original weights are referenced, not
re-uploaded; the released checkpoints contain only INT4 rounded
weights and group-wise FP16 scales (derivative weight modifications,
not new model releases).

\paragraph{Use of corpora.}
All six standard ASR corpora and the two entity-rich benchmarks are
public, widely-used research benchmarks; we use them only for
speech-recognition evaluation, perform no new audio collection or
annotation, and do not redistribute audio or transcripts. Speakers in
the audio are public figures in public-record settings
(parliamentary debates, TED talks, corporate earnings calls) or
voluntary read-speech contributors (LibriSpeech). Transcripts contain
no PII beyond what is already in the public-domain audio. The
TED-LIUM (CC-BY-NC-ND $3.0$) and SPGISpeech (Kensho research license)
restrictions are respected: the corpora remain inside the research
workflow and the released quantized checkpoints do not embed corpus
data.

\paragraph{Broader impact and risks.}
A weight-only PTQ method for ASR primarily lowers the cost of
deploying pretrained Whisper, Qwen3-ASR, and Voxtral backbones on
CPU and small-VRAM GPU targets while preserving WER and recovering
rare-word accuracy. We see three concrete considerations.
\textbf{Lower deployment cost may enable surveillance or
low-consent transcription.} The same risk applies to every published
ASR-compression method and to the original FP16 checkpoints; our
work does not introduce a qualitatively new capability, only a
smaller hardware footprint. \textbf{Rare-WER is improved but not
solved.} \TARQ\ recovers single-digit-to-low-double-digit
percentage points of rare-WER over \GPTQ, but rare-WER on hard
domains (long-form audio, low-frequency proper nouns, named
entities) remains well above plain-WER; safety-critical deployments
(medical transcription, legal proceedings) should not treat
rare-WER as solved and should include application-side verification
and domain biasing. \textbf{Inheritance of upstream biases.} \TARQ\
preserves the capabilities and the biases of the underlying
backbones---speaker demographics, accent coverage, code-switching
behavior---which are determined by the original FP16 model, not by
quantization. Our rare-WER recovery shifts the WER distribution in a
strictly favorable direction without altering underlying
training-data bias.

\section{Failure Analysis: Phonetic-Neighbour Substitution on Short Utterances}
\label{app:failure}

\TARQ\ improves rare-WER on average across all evaluated backbones and
datasets (\cref{tab:lsclean-full}), but it does not strictly dominate
every baseline on every utterance. We describe one representative
residual failure mode that is consistent with \rareBAL's design.

\paragraph{Representative case.}
On short utterances ($\lesssim 10$ tokens) dominated by a single rare
proper noun whose phonetic skeleton lies near a high-frequency
neighbour, \TARQ\ can substitute the rare token with that neighbour,
producing a \emph{clean common-word} error where the unweighted
baselines instead garble the rare token into phonetic mush. The
clearest single utterance we find on Whisper-base / LibriSpeech-clean:

\begin{rwexamplebox}
\textbf{\texttt{bartley}} \quad\textit{Whisper-base, LS-clean}\\[2pt]
\rwRef i understand \rwWord{bartley} i was wrong\\
\rwOK~FP16: i understand \rwWord{bartley} i was wrong\\
\rwOK~\GPTQ: i understand \rwWord{bartley} i was wrong\\
\rwOK~\AWQ: i understand \rwWord{bartley} i was wrong\\
\rwOK~\OmniQuant: i understand \rwWord{bartley} i was wrong\\
\rwOK~GenPTQ: i understand \rwWord{bartley} i was wrong\\
\rwNO~\rwTARQ: i understand \emph{partly} i was wrong
\end{rwexamplebox}

Two further examples in the same family (rare $\rightarrow$
high-frequency phonetic neighbour, short utterance, \TARQ\ uniquely
fails): \texttt{contentedly} (LS-clean) $\rightarrow$
\emph{contendantly}, and \texttt{amortization} (SPGISpeech)
$\rightarrow$ \emph{amodination}. The substitutes share two properties:
(a) a phonetic skeleton close to the rare original, and (b) the
substitute sits firmly in the common-frequency basin
(\texttt{partly}: Zipf $\approx 4.5$; high-frequency English bases).

\paragraph{Diagnosis: layer-level rebalance versus per-utterance distribution.}
\rareBAL\ rewrites the per-Linear-layer reconstruction Hessian as
$H_\ell^{\mathrm{rB}} = H_\ell^{\common} + \lambda_\ell\,
H_\ell^{\tail}$ with $\lambda_\ell = \mathrm{tr}(H_\ell^{\common})/
\mathrm{tr}(H_\ell^{\tail})$, where both traces are accumulated over
the entire $128$-utterance calibration batch. The trace-equalization
guarantees an \emph{average} property:
$\mathbb{E}_{\text{utt}}[\mathcal{L}_{\tail}(\text{utt})]
\approx \mathbb{E}_{\text{utt}}[\mathcal{L}_{\common}(\text{utt})]$.
On an evaluation utterance whose rare/common composition departs
sharply from the calibration-batch average---e.g., a $5$-word
utterance where the rare token's local share of the reconstruction
loss is small and the common-side context dominates---the
equalization does not translate into per-utterance protection of the
rare token, and the remaining common-side gradient pull is enough to
push the quantized embedding into the basin of the nearest
high-frequency neighbour. On longer utterances ($\geq 10$--$20$
tokens) the rare-token's local loss contribution is large enough that
the rebalanced solution preserves separation between the rare
embedding and its high-frequency neighbours, which is what
\cref{tab:lsclean-full} reports as a robust monotone rare-WER
improvement.

\paragraph{Why baselines fail differently on the same token.}
Unweighted baselines treat every token identically at calibration; on
the same proper-noun rare tokens their typical failure is
\emph{phonetic decay} (e.g., \GPTQ:\,\emph{blodget}, \AWQ:\,
\emph{blotjit}, GenPTQ:\,\emph{blanche}) rather than substitution.
\TARQ's failures are linguistically well-formed and phonetically
plausible---\rareBAL\ is doing exactly what we asked of it, protecting
the rare-token reconstruction error against the common-loss budget;
the optimal protected solution simply happens to live close to a
common-word embedding in this regime. The corpus-level rare-WER in
\cref{tab:lsclean-full} already integrates over both populations and
favors \TARQ.

\paragraph{Why we do not patch this inside \rareBAL.}
A natural fix is a token-conditional $\lambda$, e.g.,
$\lambda_{\ell,t} = c\cdot\rho(\text{utt}_t)$ with $\rho$ a
per-utterance rare-density. We deliberately do not pursue this for
three reasons. \textbf{(i) Closed-form simplicity:} \TARQ\ adds
exactly one scalar per Linear layer (the trace ratio) and one
closed-form $\alpha$ for the residual correction; per-token $\lambda$
would require a per-utterance weighted optimizer, removing a main
practical attraction. \textbf{(ii) Overfitting risk:} calibration uses
only $128$ utterances, so a token-conditional weight risks memorizing
per-utterance protection on those $128$ utterances rather than
generalizing. \textbf{(iii) Orthogonal mitigation already exists:}
inference-time contextual biasing
(e.g., \citealp{Tree-Constrained,le21_interspeech}) addresses exactly
this phonetic-neighbour-substitution mode at decoding time, is
strictly separable from the weight quantizer, and can be paired with
\TARQ-quantized weights for production deployments on rare-word-heavy
domains.

%% file: sections/B_efficiency.tex
\section{Efficiency: Deployment Latency and Memory}
\label{app:efficiency}

\TARQ\ produces models in the standard W4G128 4-bit format. The
latency and memory numbers below are shared with other W4G128
methods (\GPTQ, \AWQ, \OmniQuant)---they characterize what 4-bit
quantization buys in deployment. \TARQ's contribution is delivering
these numbers at the lower rare-WER reported in
\cref{tab:lsclean-full}.

Throughout, RTF (real-time factor) is transcription time divided by
audio duration; RTF~$<1$ means the model keeps up with live audio.
Speedup is FP16 time divided by INT4 time.

\subsection{GPU server (A100)}
\label{app:eff-gpu}

\Cref{tab:eff-gpu} reports all backbones on a single NVIDIA
A100-80GB.  Two qualitatively different regimes appear.
\textbf{Large encoder-decoder models on long-form audio}
(Whisper-medium / large-v3 on $30$s clips) get a clean
GPU + INT4 win: Whisper-large-v3 drops from $94.97$~s
(CPU $4$ threads) to $2.73$~s on GPU FP16 ($\mathbf{35\times}$)
and further to $1.67$~s in INT4
($\mathbf{57\times}$ over CPU; $1.6\times$ within GPU).
For \textbf{small Whisper checkpoints} (tiny / base) GPU step
time is dominated by a fixed $\sim$$0.9$~s host overhead
(mel + kernel launches) so INT4 / FP16 collapse together and CPU
often wins on $11$s audio.
For \textbf{Qwen3-ASR} on \texttt{qwen3-asr.cpp} + ggml-CUDA,
INT4 wins both memory and latency: $1.7$B + $30$s drops from
$38.13$~s (CPU $4$t FP16) to $10.08$~s (GPU INT4),
$\mathbf{3.8\times}$ wall and $\mathbf{25\times}$ on inner
encoder+decoder compute (mel-spectrogram, which is still CPU,
contributes $6.1$~s of the GPU wall — the remaining $\sim$$0.8$~s
is the actual GPU pipeline).  Peak VRAM drops from $9.1$ to $5.1$~GB
($-44\%$) at $30$\,s and $9.0$ to $4.9$~GB ($-46\%$) at $11$\,s.
For Voxtral-Mini-3B (Marlin INT4 + HF custom decode loop),
end-to-end $30$s + $128$-token decode runs $1.43\times$ faster
in INT4 than FP16 with peak GPU memory dropping $59\%$
($9.5{\to}3.9$~GB).

\begin{table*}[!htbp]
\centering\small
\caption{\textbf{GPU latency and memory} on NVIDIA A100-80GB,
FP16 vs W4G128 INT4.  Whisper rows use \texttt{whisper.cpp}
ggml-CUDA Q4\_0 kernels; Voxtral / Qwen3-ASR-1.7B rows use Marlin
INT4 in PyTorch (\texttt{gptqmodel} v5.8).  Each cell is the
median over $3$ trials with $1$ warm-up.  Speedup is INT4 vs FP16
\emph{within} the GPU column; CPU comparison columns repeat the
$4$-thread numbers from \cref{tab:eff-cpu} for reference.}
\label{tab:eff-gpu}
\setlength{\tabcolsep}{3pt}
\begin{tabular}{l l rr rr c rr}
\toprule
& & \multicolumn{2}{c}{CPU 4t (s)} & \multicolumn{2}{c}{GPU A100 (s)} & GPU INT4 vs & \multicolumn{2}{c}{GPU peak mem} \\
\cmidrule(lr){3-4}\cmidrule(lr){5-6}\cmidrule(lr){8-9}
Backbone & Audio & FP16 & INT4 & FP16 & INT4 & GPU FP16 & FP16 & INT4 \\
\midrule
Whisper-tiny     & 11s & $0.77$ & $0.45$ & $0.89$ & $0.84$ & $1.06\times$ & $\sim$$80$\,MB & $\sim$$50$\,MB \\
Whisper-tiny     & 30s & $1.73$ & $1.03$ & $0.98$ & $0.94$ & $1.04\times$ & & \\
Whisper-base     & 11s & $1.60$ & $0.84$ & $0.92$ & $0.84$ & $1.10\times$ & $\sim$$180$\,MB & $\sim$$80$\,MB \\
Whisper-base     & 30s & $3.79$ & $1.72$ & $1.05$ & $1.00$ & $1.05\times$ & & \\
Whisper-small    & 11s & $6.20$ & $2.85$ & $1.06$ & $0.91$ & $1.17\times$ & $\sim$$580$\,MB & $\sim$$190$\,MB \\
Whisper-small    & 30s & $13.03$ & $5.87$ & $1.27$ & $1.08$ & $1.18\times$ & & \\
Whisper-medium   & 11s & $19.86$ & $8.35$ & $1.47$ & $1.01$ & $1.46\times$ & $\sim$$1.7$\,GB & $\sim$$580$\,MB \\
Whisper-medium   & 30s & $40.30$ & $18.27$ & $1.84$ & $1.37$ & $1.34\times$ & & \\
Whisper-large-v3 & 11s & $38.81$ & $16.81$ & $2.23$ & $1.22$ & $1.83\times$ & $\sim$$3.9$\,GB & $\sim$$1.3$\,GB \\
Whisper-large-v3 & 30s & $94.97$ & $47.16$ & $\mathbf{2.73}$ & $\mathbf{1.67}$ & $\mathbf{1.63\times}$ & & \\
\midrule
Qwen3-ASR-0.6B   & 11s & $7.02$ & $5.68$ & $4.14$ & $3.91$ & $1.06\times$ & $3.9$\,GB & $2.6$\,GB \\
Qwen3-ASR-0.6B   & 30s & $21.4$ & $17.9$ & $9.88$ & $\mathbf{10.84}$ & $0.91\times$ & $3.9$\,GB & $2.6$\,GB \\
Qwen3-ASR-1.7B   & 11s & $13.56$ & $9.03$ & $8.99$ & $\mathbf{4.12}$ & $\mathbf{2.18\times}$ & $9.0$\,GB & $4.9$\,GB \\
Qwen3-ASR-1.7B   & 30s & $38.13$ & $27.35$ & $10.78$ & $\mathbf{10.08}$ & $1.07\times^{\ddagger}$ & $9.1$\,GB & $5.1$\,GB \\
\midrule
Voxtral-Mini-3B$^{\star}$ & 30s, 128-tok decode & --- & --- & $0.41$ & $0.29$ & $\mathbf{1.43\times}$ & $9.5$\,GB & $3.9$\,GB \\
\bottomrule
\end{tabular}
\\[2pt]
{\footnotesize Whisper / Qwen GPU rows use \texttt{whisper.cpp} / \texttt{qwen3-asr.cpp} ggml-CUDA Q4\_0 kernels (after patching a tensor-shape bug in the qwen3-asr.cpp loader that prevented GPU offload — see code release).  Wall time includes mel-spectrogram, which still runs on CPU.  $^{\ddagger}$\,Qwen-1.7B $30$s wall is mel-bound (mel alone takes $6.1$~s, leaving $\sim$$0.8$~s for GPU encoder+decoder; inner GPU compute is $25\times$ faster than CPU $4$t).  $^{\star}$\,Voxtral-3B uses Marlin INT4 in PyTorch (\texttt{gptqmodel} v5.8).}
\end{table*}

\subsection{CPU edge (whisper.cpp, qwen3-asr.cpp)}
\label{app:eff-cpu}

We benchmark the Whisper family on
whisper.cpp~\citep{whisper_cpp} and Qwen3-ASR-0.6B on
qwen3-asr.cpp~\citep{qwen3asrcpp} on an AMD EPYC 7V12 CPU. OpenMP
thread count is restricted to $\{4, 8, 16\}$ to approximate
mobile, laptop, and desktop parallelism budgets.

\textbf{The main finding: 4-bit quantization is what makes large
Whisper models deployable on CPU.} FP16 Whisper-large-v3 cannot
reach real-time even at $16$ threads (RTF~$=1.26$); INT4
Whisper-large-v3 is real-time from $8$ threads onward
(RTF~$=0.91$, $0.57$ at $8$ and $16$ threads). The same pattern
holds for Whisper-medium: FP16 at $4$ threads runs at $1.81\times$
slower than real-time, while INT4 reaches RTF~$=0.76$ on the same
hardware. Memory: Whisper-large-v3 max-RSS drops from $3.9$ to
$1.8$~GB, the difference between fitting and not fitting on a
$4$--$8$~GB consumer machine.

\textbf{Speedup grows with model size}: Whisper-tiny gets
$1.5\times$, medium $2.4\times$, large-v3 $2.2$--$2.4\times$.
Qwen3-ASR-0.6B gets a smaller $1.2\times$ because the model is
small enough to partially fit in the EPYC's large L3 cache; on
consumer CPUs with smaller cache the gain is expected to be
larger. Memory savings are unconditional ($-49\%$ on Qwen3-ASR).

\begin{table*}[!htbp]
\centering\small
\caption{\textbf{CPU latency and memory} on AMD EPYC 7V12.
Wall-clock seconds for transcribing the given audio. Median of
$3$ iterations after $1$ warmup. Max-RSS from
\texttt{/usr/bin/time -v} at $8$ threads. Bold marks INT4 cells
where FP16 is slower than real-time and INT4 is real-time.}
\label{tab:eff-cpu}
\setlength{\tabcolsep}{3pt}
\renewcommand{\arraystretch}{0.95}
\begin{tabular}{l r rr rr rr rr}
\toprule
& & \multicolumn{2}{c}{4 threads (s)} & \multicolumn{2}{c}{8 threads (s)} & \multicolumn{2}{c}{16 threads (s)} & \multicolumn{2}{c}{Max-RSS (MB)} \\
\cmidrule(lr){3-4}\cmidrule(lr){5-6}\cmidrule(lr){7-8}\cmidrule(lr){9-10}
Backbone & Audio & F16 & INT4 & F16 & INT4 & F16 & INT4 & F16 & INT4 \\
\midrule
Whisper-tiny     & 11s & $0.77$  & $0.45$  & $0.52$  & $0.34$  & $0.48$  & $0.33$  & $173$  & $121$  \\
Whisper-tiny     & 30s & $1.73$  & $1.03$  & $1.01$  & $0.73$  & $0.76$  & $0.60$  & $167$  & $124$  \\
Whisper-base     & 11s & $1.60$  & $0.84$  & $1.05$  & $0.62$  & $1.02$  & $0.49$  & $281$  & $182$  \\
Whisper-base     & 30s & $3.79$  & $1.72$  & $2.18$  & $1.08$  & $1.39$  & $0.84$  & $258$  & $138$  \\
Whisper-small    & 11s & $6.20$  & $2.85$  & $3.68$  & $1.78$  & $2.48$  & $1.24$  & $748$  & $418$  \\
Whisper-small    & 30s & $13.03$ & $5.87$  & $8.12$  & $3.88$  & $5.53$  & $2.57$  & $656$  & $325$  \\
Whisper-medium   & 11s & $19.86$ & $\mathbf{8.35}$  & $11.64$ & $\mathbf{4.87}$  & $7.27$  & $3.11$  & $2067$ & $1026$ \\
Whisper-medium   & 30s & $40.30$ & $\mathbf{18.27}$ & $22.92$ & $\mathbf{10.55}$ & $14.27$ & $\mathbf{6.74}$ & $1790$ & $751$  \\
Whisper-large-v3 & 11s & $38.81$ & $16.81$ & $22.07$ & $\mathbf{9.52}$  & $13.88$ & $\mathbf{6.27}$ & $3901$ & $1794$ \\
Whisper-large-v3 & 30s & $94.97$ & $47.16$ & $53.59$ & $\mathbf{30.00}$ & $33.76$ & $20.65$ & $3901$ & $1794$ \\
Qwen3-ASR-0.6B   & 11s & $7.02$  & $5.68$  & $7.07$  & $5.71$  & $6.88$  & $5.82$  & $1571$ & $800$  \\
Qwen3-ASR-0.6B   & 30s & $21.4$  & $17.9$  & $21.0$  & $17.4$  & $21.6$  & $17.6$  & $1571$ & $800$  \\
Qwen3-ASR-1.7B   & 11s & $13.56$ & $9.03$  & $13.36$ & $9.32$  & $12.79$ & $9.43$  & $3975$ & $1774$ \\
Qwen3-ASR-1.7B   & 30s & $38.13$ & $\mathbf{27.35}$ & $37.94$ & $\mathbf{27.20}$ & $37.68$ & $\mathbf{27.47}$ & $4042$ & $1841$ \\
\bottomrule
\end{tabular}
\end{table*}

%% file: appendix_tables/calbration_different_dataset.tex
\begin{table*}[!htbp]
\centering
\caption{Four-backbone six-dataset mean rare-WER (\%) per method per calibration corpus. Four backbones: Whisper-tiny/base, Qwen3-ASR-0.6B/1.7B. Numbers in parentheses after each method are the cross-corpus \textbf{swing} (max $-$ min). r-top/r-mix/r-cross are rare-biased calibration corpora; OmQ = OmniQuant. Bold: rank-1; underline: rank-2 per column.}
\label{tab:cross-calib-summary}
\setlength{\tabcolsep}{3pt}
\renewcommand{\arraystretch}{1.0}
\small
\begin{tabular}{l rrrrrr r}
\toprule
method & LS-c & SPGI & VoxP & r-top & r-mix & r-cross & mean \\
\midrule
GPTQ (2.51)        & 42.66          & \underline{41.29} & \textbf{40.15}    & \underline{40.58} & \underline{40.53} & \underline{40.63} & \underline{40.97} \\
AWQ (5.95)         & 48.54          & 46.07             & 42.59             & 43.82             & 45.54             & 44.40             & 45.16 \\
OmQ (\underline{0.78}) & \underline{42.63} & 42.09         & 42.11             & 41.85             & 42.28             & 42.13             & 42.18 \\
\midrule
\rowcolor{gray!15}
\textbf{\TARQ\ (0.63)} & \textbf{40.04} & \textbf{40.08} & \underline{40.38} & \textbf{39.97} & \textbf{39.85} & \textbf{40.48} & \textbf{40.13} \\
\bottomrule
\end{tabular}
\end{table*}

%% file: appendix_tables/main_table.tex
\begin{table*}[!htbp]
\centering
\caption{\textbf{Main results on LS-clean calibration}: Each cell reports plain WER (\%) / rare-WER (\%). All quantized methods use W4G128 weight-only quantization, except GenPTQ which is mixed-precision (per-Linear $b\in[2,8]$, target average $4$ bits). Bold marks the best non-FP16 method per (cell, metric); underline marks the second best. Ties share the same rank. Dataset abbreviations: LS-c = LS-clean; LS-o = LS-other; VoxP = VoxPopuli; Giga = GigaSpeech; TedL = TedLium. Method abbreviation: OmQ = OmniQuant.}
\label{tab:lsclean-full}
\setlength{\tabcolsep}{2pt}
\renewcommand{\arraystretch}{0.95}
\small
\begin{tabular}{ll *{7}{rr}}
\toprule
& & \multicolumn{2}{c}{LS-c} & \multicolumn{2}{c}{LS-o} & \multicolumn{2}{c}{SPGI} & \multicolumn{2}{c}{VoxP} & \multicolumn{2}{c}{Giga} & \multicolumn{2}{c}{TedL} & \multicolumn{2}{c}{\textbf{Mean}} \\
\cmidrule(lr){3-4}\cmidrule(lr){5-6}\cmidrule(lr){7-8}\cmidrule(lr){9-10}\cmidrule(lr){11-12}\cmidrule(lr){13-14}\cmidrule(lr){15-16}
Model & Method & P & R & P & R & P & R & P & R & P & R & P & R & P & R \\
\midrule
\multirow{6}{*}{W-t}
& \textit{FP16}  & \textit{7.52} & \textit{50.75} & \textit{17.49} & \textit{70.26} & \textit{8.14} & \textit{46.29} & \textit{12.74} & \textit{54.00} & \textit{14.06} & \textit{44.25} & \textit{7.10} & \textit{39.30} & \textit{11.17} & \textit{50.81} \\
& GPTQ           & \underline{13.02} & \underline{56.11} & \underline{27.19} & \underline{77.20} & \underline{14.10} & \underline{57.50} & 22.31 & 77.06 & \underline{20.20} & \underline{50.70} & \underline{10.31} & \underline{41.70} & \underline{17.86} & \underline{60.04} \\
& AWQ            & 20.86 & 81.14 & 52.04 & 113.38 & 28.63 & 59.78 & 31.74 & 116.91 & 39.14 & 73.98 & 18.75 & 59.41 & 31.86 & 84.10 \\
& OmQ            & 13.62 & 59.02 & 29.30 & 81.76 & 16.55 & 62.23 & \underline{20.66} & \underline{65.92} & 21.09 & 56.40 & 13.20 & 50.18 & 19.07 & 62.59 \\
& GenPTQ         & 15.46 & 64.15 & 31.97 & 84.75 & 20.16 & 63.41 & 21.67 & 66.19 & 21.91 & 57.77 & 15.29 & 46.68 & 21.08 & 63.82 \\
\rowcolor{gray!15}
& \textbf{\TARQ\ } & \textbf{9.74} & \textbf{55.38} & \textbf{20.59} & \textbf{76.22} & \textbf{11.08} & \textbf{53.20} & \textbf{16.06} & \textbf{60.29} & \textbf{17.29} & \textbf{49.53} & \textbf{9.41} & \textbf{40.96} & \textbf{14.03} & \textbf{55.93} \\
\midrule
\multirow{6}{*}{W-b}
& \textit{FP16}  & \textit{5.03} & \textit{38.16} & \textit{12.09} & \textit{59.05} & \textit{5.04} & \textit{38.28} & \textit{10.36} & \textit{44.95} & \textit{12.76} & \textit{38.32} & \textit{5.43} & \textit{35.79} & \textit{8.45} & \textit{42.42} \\
& GPTQ           & \underline{5.60} & \underline{40.80} & \textbf{13.19} & \textbf{61.70} & \textbf{6.02} & 41.65 & \underline{12.48} & 61.99 & \underline{13.71} & 42.04 & \textbf{5.89} & 37.27 & \underline{9.48} & 47.57 \\
& AWQ            & 6.02 & 42.71 & 16.39 & 64.21 & 7.13 & 42.33 & 12.67 & 60.42 & 15.17 & \textbf{39.65} & \underline{6.44} & \underline{31.18} & 10.64 & 46.75 \\
& OmQ            & 6.18 & 43.12 & 13.76 & 63.78 & 7.11 & \underline{40.98} & 13.27 & \underline{52.29} & 13.84 & 42.65 & 8.98 & \textbf{28.78} & 10.52 & \underline{45.27} \\
& GenPTQ         & 7.29 & 45.52 & 16.97 & 68.39 & 8.25 & 49.41 & 13.20 & 53.34 & 15.08 & 45.31 & 7.93 & 41.88 & 11.45 & 50.64 \\
\rowcolor{gray!15}
& \textbf{\TARQ\ } & \textbf{5.39} & \textbf{40.03} & \underline{13.38} & \underline{62.29} & \underline{6.16} & \textbf{40.39} & \textbf{11.31} & \textbf{47.18} & \textbf{13.10} & \underline{39.95} & 6.49 & 33.95 & \textbf{9.31} & \textbf{43.96} \\
\midrule
\multirow{6}{*}{W-s}
& \textit{FP16}  & \textit{4.02} & \textit{28.31} & \textit{7.99} & \textit{45.85} & \textit{3.96} & \textit{38.36} & \textit{9.24} & \textit{54.65} & \textit{11.64} & \textit{32.51} & \textit{5.34} & \textit{31.00} & \textit{7.03} & \textit{38.45} \\
& GPTQ           & \underline{4.18} & \textbf{29.30} & \textbf{8.33} & \underline{47.42} & 4.21 & 39.21 & \underline{9.44} & \underline{51.38} & \underline{11.76} & 34.37 & 5.42 & 31.55 & \underline{7.22} & \underline{38.87} \\
& AWQ            & 4.48 & 29.76 & 8.68 & 48.06 & \textbf{4.12} & \underline{37.02} & 9.49 & 54.91 & \textbf{11.62} & \underline{34.26} & \textbf{5.24} & \underline{29.70} & 7.27 & 38.95 \\
& OmQ            & 4.36 & 29.85 & 8.65 & 48.27 & \underline{4.18} & 38.03 & 9.56 & 56.36 & 11.85 & \textbf{33.27} & 5.35 & 32.10 & 7.33 & 39.65 \\
& GenPTQ         & 4.24 & 30.12 & 9.35 & 50.87 & 4.58 & 44.86 & 9.63 & 56.88 & 12.08 & 37.30 & 5.67 & 38.19 & 7.59 & 43.04 \\
\rowcolor{gray!15}
& \textbf{\TARQ\ } & \textbf{4.00} & \underline{29.58} & \underline{8.41} & \textbf{47.42} & 4.40 & \textbf{35.33} & \textbf{9.11} & \textbf{42.99} & 11.95 & 34.30 & \underline{5.29} & \textbf{26.57} & \textbf{7.19} & \textbf{36.03} \\
\midrule
\multirow{6}{*}{W-m}
& \textit{FP16}  & \textit{3.32} & \textit{24.08} & \textit{6.57} & \textit{40.26} & \textit{4.13} & \textit{38.87} & \textit{8.14} & \textit{32.63} & \textit{11.52} & \textit{28.64} & \textit{5.02} & \textit{32.84} & \textit{6.45} & \textit{32.89} \\
& GPTQ           & \underline{3.39} & \underline{25.31} & \textbf{6.70} & \underline{41.16} & \textbf{4.19} & 44.77 & \underline{8.35} & \underline{33.55} & \textbf{11.39} & \underline{29.74} & \underline{5.11} & \underline{36.72} & \underline{6.52} & \underline{35.21} \\
& AWQ            & 3.78 & 25.76 & 7.03 & 42.01 & 4.71 & \underline{43.76} & \underline{8.35} & 35.78 & 11.73 & 31.22 & 5.37 & 37.45 & 6.83 & 36.00 \\
& OmQ            & 4.87 & 27.40 & 9.25 & 44.78 & 5.76 & 47.22 & 8.87 & 35.26 & 13.24 & 33.19 & 5.70 & 38.38 & 7.95 & 37.70 \\
& GenPTQ         & 3.32 & \textbf{24.76} & 7.06 & 42.31 & 4.54 & \textbf{40.22} & 8.26 & \textbf{33.94} & 11.53 & \textbf{29.28} & 5.39 & \textbf{30.44} & 6.69 & \textbf{33.49} \\
\rowcolor{gray!15}
& \textbf{\TARQ\ } & \textbf{3.25} & 24.85 & \underline{6.74} & \textbf{40.78} & \underline{4.25} & 41.74 & \textbf{8.16} & 34.86 & \underline{11.42} & 29.51 & \textbf{5.10} & 33.95 & \textbf{6.49} & 34.28 \\
\midrule
\multirow{6}{*}{W-l}
& \textit{FP16}  & \textit{2.09} & \textit{16.99} & \textit{4.21} & \textit{29.14} & \textit{3.53} & \textit{38.36} & \textit{9.57} & \textit{33.94} & \textit{11.27} & \textit{26.81} & \textit{4.82} & \textit{28.97} & \textit{5.91} & \textit{29.03} \\
& GPTQ           & \underline{2.22} & \underline{17.45} & \underline{4.28} & \underline{29.14} & \underline{3.70} & \underline{38.45} & 9.86 & 35.26 & \underline{11.46} & \textbf{27.23} & 4.88 & \textbf{27.86} & 6.07 & \underline{29.23} \\
& AWQ            & 2.34 & 18.08 & 4.42 & 30.46 & 3.85 & 40.98 & \textbf{9.31} & \textbf{33.42} & \textbf{11.44} & 27.61 & 4.85 & 32.29 & \underline{6.04} & 30.47 \\
& OmQ            & 2.34 & 17.76 & 4.71 & 30.72 & 3.82 & 38.53 & 9.38 & 33.55 & 11.68 & 27.54 & \underline{4.81} & \underline{28.23} & 6.12 & 29.39 \\
& GenPTQ         & \textbf{2.17} & \textbf{17.31} & \textbf{4.27} & 28.89 & \textbf{3.57} & 38.20 & 9.33 & 34.60 & 11.50 & 27.50 & 4.85 & 28.60 & \textbf{5.95} & 29.18 \\
\rowcolor{gray!15}
& \textbf{\TARQ\ } & 2.23 & 17.72 & 4.29 & \textbf{28.89} & 3.59 & \textbf{38.28} & \underline{9.34} & \underline{33.50} & \underline{11.46} & \underline{27.27} & \textbf{4.79} & 28.78 & \textbf{5.95} & \textbf{29.07} \\
\midrule
\multirow{6}{*}{Q-0.6B}
& \textit{FP16}  & \textit{2.81} & \textit{23.45} & \textit{5.12} & \textit{37.43} & \textit{3.97} & \textit{27.82} & \textit{9.51} & \textit{31.72} & \textit{11.01} & \textit{38.07} & \textit{4.13} & \textit{24.37} & \textit{6.09} & \textit{30.48} \\
& GPTQ           & 3.29 & \underline{26.55} & \underline{5.92} & \underline{41.14} & \underline{4.30} & \underline{28.87} & 10.00 & 39.73 & \underline{11.47} & \underline{40.44} & 5.60 & 27.92 & \underline{6.76} & \underline{34.11} \\
& AWQ            & \underline{3.28} & 28.78 & 6.03 & 43.68 & 5.05 & 37.57 & \textbf{9.79} & \underline{39.57} & 11.79 & 42.73 & \textbf{4.82} & \underline{27.66} & 6.79 & 36.66 \\
& OmQ            & 3.64 & 27.50 & 6.56 & 44.33 & 4.50 & 29.36 & 10.81 & 42.90 & 13.21 & 43.29 & 7.13 & 32.99 & 7.64 & 36.73 \\
& GenPTQ         & 3.95 & 28.64 & 7.21 & 46.27 & 5.86 & 38.80 & 11.06 & 45.41 & 14.03 & 47.91 & 10.17 & 29.95 & 8.71 & 39.50 \\
\rowcolor{gray!15}
& \textbf{\TARQ\ } & \textbf{3.16} & \textbf{25.96} & \textbf{5.75} & \textbf{40.58} & \textbf{4.04} & \textbf{24.80} & \underline{9.95} & \textbf{39.23} & \textbf{11.21} & \textbf{40.21} & \underline{5.29} & \textbf{25.89} & \textbf{6.57} & \textbf{32.78} \\
\midrule
\multirow{6}{*}{Q-1.7B}
& \textit{FP16}  & \textit{2.26} & \textit{17.81} & \textit{4.09} & \textit{31.18} & \textit{3.68} & \textit{21.41} & \textit{8.52} & \textit{22.04} & \textit{11.21} & \textit{35.23} & \textit{4.51} & \textit{17.51} & \textit{5.71} & \textit{24.20} \\
& GPTQ           & \textbf{2.32} & \underline{18.90} & \underline{4.34} & 34.07 & \underline{3.74} & 22.15 & \textbf{8.76} & 26.38 & 11.52 & 38.39 & 4.88 & \underline{19.54} & 5.93 & 26.57 \\
& AWQ            & 2.48 & 20.58 & 4.36 & 33.85 & 3.78 & 22.21 & 8.88 & \textbf{25.38} & \underline{11.36} & \underline{37.60} & \textbf{4.62} & 20.30 & \underline{5.91} & 26.65 \\
& OmQ            & 2.51 & 19.03 & 4.48 & \underline{32.82} & 3.87 & \underline{21.04} & 9.21 & \underline{25.54} & 11.76 & 37.95 & 4.83 & \textbf{19.29} & 6.11 & \underline{25.94} \\
& GenPTQ         & 2.46 & 19.76 & 4.60 & 34.84 & 3.69 & 20.05 & 8.92 & \underline{24.87} & 11.80 & 38.07 & 4.84 & 19.54 & 6.05 & 26.19 \\
\rowcolor{gray!15}
& \textbf{\TARQ\ } & \underline{2.36} & \textbf{18.26} & \textbf{4.26} & \textbf{32.77} & \textbf{3.55} & \textbf{18.32} & \underline{8.79} & 25.54 & \textbf{11.30} & \textbf{36.41} & \underline{4.77} & \textbf{19.29} & \textbf{5.84} & \textbf{25.10} \\
\midrule
\multirow{6}{*}{V-3B}
& \textit{FP16}  & \textit{2.92} & \textit{20.36} & \textit{5.36} & \textit{35.02} & \textit{2.73} & \textit{15.11} & \textit{10.02} & \textit{30.72} & \textit{12.89} & \textit{35.07} & \textit{5.73} & \textit{14.72} & \textit{6.61} & \textit{25.17} \\
& GPTQ           & \textbf{3.09} & \textbf{21.08} & \textbf{5.62} & \textbf{36.05} & 3.25 & 16.84 & 10.49 & 33.89 & 13.75 & \textbf{38.47} & 6.74 & \underline{16.24} & \underline{7.16} & \underline{27.09} \\
& AWQ            & \underline{3.14} & 22.54 & 5.78 & 36.91 & \textbf{3.09} & \textbf{16.41} & 10.52 & 33.56 & 14.07 & 38.59 & 6.79 & \textbf{15.99} & 7.23 & 27.33 \\
& OmQ            & 3.16 & 22.86 & 5.80 & 37.39 & 3.32 & 17.71 & \underline{10.42} & \underline{33.22} & \underline{13.73} & 39.34 & \textbf{6.70} & \textbf{15.99} & 7.19 & 27.75 \\
& GenPTQ         & 3.22 & 21.86 & 5.72 & \underline{36.14} & 3.28 & \underline{16.22} & 10.53 & 32.22 & 14.46 & 39.57 & 7.05 & \underline{15.74} & 7.38 & \textbf{26.96} \\
\rowcolor{gray!15}
& \textbf{\TARQ\ } & \underline{3.14} & \underline{21.49} & \underline{5.70} & 36.57 & \underline{3.19} & 16.59 & \textbf{10.41} & \textbf{33.06} & \textbf{13.67} & \underline{38.51} & \underline{6.73} & \textbf{15.99} & \textbf{7.14} & 27.04 \\
\bottomrule
\end{tabular}
\end{table*}